\documentclass[runningheads, envcountsame, a4paper]{llncs}
\usepackage{graphicx}
\usepackage[misc]{ifsym}
\usepackage{array,makecell,rotating}
\usepackage{subcaption}
\usepackage{algorithm}
\usepackage{algpseudocode}
\usepackage{multirow} 
\usepackage{ulem}
\usepackage{stix}
\usepackage{amsmath,amssymb,amsfonts}
\usepackage{cite}
\usepackage[hyphens]{url}

\begin{document}
\title{A Stochastic Quasi-Newton Method with Nesterov's Accelerated Gradient}
\toctitle{A Stochastic Quasi-Newton Method with Nesterov's Accelerated Gradient}

%\thanks{Supported by organization x.}

\author{S. Indrapriyadarsini\inst{1}{\Letter} \and
Shahrzad Mahboubi\inst{2}\and
Hiroshi Ninomiya\inst{2}\and
Hideki Asai\inst{1}}
\authorrunning{S. Indrapriyadarsini et al.}

\tocauthor{S. Indrapriyadarsini, Shahrzad Mahboubi, Hiroshi Ninomiya and Hideki Asai}

%
%\institute{Princeton University, Princeton NJ 08544, USA \and
%Springer Heidelberg, Tiergartenstr. 17, 69121 Heidelberg, Germany

%
\institute{Shizuoka University, Hamamatsu, Shizuoka Pre., Japan 
\email{$\{$s.indrapriyadarsini.17,asai.hideki$\}$@shizuoka.ac.jp}\and
Shonan Institute of Technology, Fujisawa, Kanagawa Pre., Japan
\email{$\{$18T2012@sit,ninomiya@info$\}$.shonan-it.ac.jp}}
%\email{s.indrapriyadarsini.17@shizuoka.ac.jp}\\
%\email{18T2012@sit.shonan-it.ac.jp}\\
%\email{ninomiya@info.shonan-it.ac.jp}\\
%\email{asai.hideki@shizuoka.ac.jp}\\
%
\maketitle              % typeset the header of the contribution
\begin{abstract}
Incorporating second order curvature information in gradient based methods have shown to improve convergence drastically despite its computational intensity. 
%Further, studies on stochastic adaptations of the quasi-Newton methods in limited memory have gained interests. 
In this paper, we propose a stochastic (online) quasi-Newton method with Nesterov's accelerated gradient in both its full and limited memory forms for solving large scale non-convex optimization problems in neural networks. 
%Direction normalization has been introduced to improve stabilty.
%Further we also include direction normalization to improve stability. %
The performance of the proposed algorithm is evaluated in Tensorflow on benchmark classification and regression problems. The results show  improved performance compared to the classical second order oBFGS and oLBFGS methods and popular first order stochastic methods such as SGD and Adam. The performance with different momentum rates and batch sizes have also been illustrated.
%The abstract should briefly summarize the contents of the paper in

%Stochastic optimization, Nesterov's accelerated gradient, quasi-Newton method, limited memory, Tensorflow 
\end{abstract}
\section{Introduction}
Neural networks have shown to be effective in innumerous real-world applications. Most of these applications require large neural network models with massive amounts of training data to achieve good accuracies and low errors. Neural network optimization poses several challenges such as ill-conditioning, vanishing and exploding gradients, choice of hyperparameters, etc. Thus choice of the optimization algorithm employed on the neural network model plays an important role. It is expected that the neural network training imposes relatively lower computational and memory demands, in which case a full-batch approach is not suitable. Thus, in large scale optimization problems, a stochastic approach is more desirable.
Stochastic optimization algorithms use a small subset of data (mini-batch) in its evaluations of the objective function. These methods are particularly of relevance in examples of a continuous stream of data, where the partial data is to be modelled as it arrives. Since the stochastic or online methods operate on small subsamples of the data and its gradients, they significantly reduce the computational and memory requirements.% ~\cite{schraudolph2007stochastic}

\subsection{Related Works}
Gradient based algorithms are popularly used in training neural network models. These algorithms can be broadly classified into first order and second order methods \cite{haykin2009neural}. Several works have been devoted to stochastic first-order methods such as stochastic gradient descent (SGD) \cite{bottou2004large,bottou2010large} and its variance-reduced forms \cite{robbins1951stochastic, peng2019accelerating , johnson2013accelerating}, AdaGrad \cite{duchi2011adaptive}, RMSprop \cite{tieleman2012lecture} and Adam \cite{kingma2014adam}. First order methods are popular due to its simplicity and optimal complexity. However, incorporating the second order curvature information have shown to improve convergence. But one of the major drawbacks in second order methods is its need for high computational and memory resources. Thus several approximations have been proposed under Newton\cite{martens2010deep, roosta2016sub} and quasi-Newton\cite{dennis1977quasi} %~\cite{ninomiya2017novel, byrd2016stochastic, mokhtari2015global, schraudolph2007stochastic} 
methods in order to make use of the second order information while keeping the computational load minimal.  

Unlike the first order methods, getting quasi-Newton methods to work in a stochastic setting is challenging and has been an active area of research. The oBFGS method \cite{schraudolph2007stochastic} is one of the early stable stochastic quasi-Newton methods, in which the gradients are computed twice using the same sub-sample, to ensure stability and scalability. Recently there has been a surge of interest in designing efficient stochastic second order variants which are better suited for large scale problems. \cite{mokhtari2014res} proposed a regularized stochastic BFGS method (RES) that modifies the proximity condition of BFGS.  \cite{mokhtari2015global} further analyzed the global convergence properties of stochastic BFGS and proposed an online L-BFGS method.  \cite{byrd2016stochastic} proposed a stochastic limited memory BFGS (SQN) through sub-sampled Hessian vector products. \cite{wang2017stochastic} proposed a general framework for stochastic quasi-Newton methods that assume noisy gradient information through first order oracle (SFO) and extended it to a stochastic damped L-BFGS method (SdLBFGS). This was further modified in \cite{li2018implementation} by reinitializing the Hessian matrix at each iteration to improve convergence and normalizing the search direction to improve stability. There are also several other studies on stochastic quasi-Newton methods with variance reduction~\cite{lucchi2015variance, moritz2016linearly, bollapragada2018progressive}, sub-sampling\cite{byrd2011use, roosta2016sub} and block updates\cite{gower2016stochastic}. Most of these methods have been proposed for solving convex optimization problems, but training of neural networks for non-convex problems have not been mentioned in their scopes. The focus of this paper is on training neural networks for non-convex problems with methods similar to that of the oBFGS in \cite{schraudolph2007stochastic} and RES \cite{mokhtari2014res,mokhtari2015global}, as they are stochastic extensions of the classical quasi-Newton method. Thus, the other sophisticated algorithms \cite{byrd2016stochastic, wang2017stochastic, li2018implementation, lucchi2015variance, moritz2016linearly, bollapragada2018progressive, byrd2011use, roosta2016sub, gower2016stochastic} are excluded from comparision in this paper and will be studied in future works.

In this paper, we introduce a novel stochastic quasi-Newton method that is accelerated using Nesterov's accelerated gradient. Acceleration of quasi-Newton method with Nesterov's accelerated gradient have shown to improve convergence \cite{ninomiya2017novel,LNAQ_shah}. The proposed algorithm is a stochastic extension of the accelerated methods in \cite{ninomiya2017novel,LNAQ_shah} with changes similar to the oBFGS method. The proposed method is also discussed both in its full and limited memory forms. The performance of the proposed methods are evaluated on benchmark classification and regression problems and compared with the conventional SGD, Adam and o(L)BFGS methods.
%\cite{schraudolph2007stochastic,mokhtari2015global,byrd2016stochastic}, RES\cite{mokhtari2014res}, SDBFGS\cite{wang2017stochastic,li2018implementation}, SFO\cite{sohl2013fast}, SQN\cite{byrd2016stochastic} 

%Some of these include \cite{byrd2011use}  Several variants of the stochastic quasi-Newton method~\cite{li2018implementation,schraudolph2007stochastic,byrd2016stochastic} have been proposed. By making descent steps which are scaled by the approximate inverse Hessian, and which are therefore longer in directions of shallow curvature and shorter in directions of steep curvature, quasi-Newton methods can be orders of magnitude faster than steepest descent \cite{sohl2013fast}
%\vspace{-1.4mm}
\section{Background}
%\vspace{-1.4mm}
\begin{equation}
\label{eq:obj} 
\underset{{\bf w} \in \mathbb{R}^d}{\text {min}} E({\bf w})= \frac {1}{b}{ \sum_{p \in X} E_p ({\bf w})}, 
\end{equation} 

%\section{Stochastic Gradient Based Methods}
Training in neural networks is an iterative process in which the parameters are updated in order to minimize an objective function. Given a mini-batch ${X \subseteq T_r}$  with samples ${(x_p,d_p)_{p \in X}}$ drawn at random from the training set ${T_r}$ and error function ${E_p({\bf w} ;x_p,d_p)}$ parameterized by a vector ${{\bf w} \in \mathbb{R}^d}$,  the objective function is defined as in (\ref{eq:obj})
where ${b}={|X|}$, is the batch size. In full batch, ${X = T_r }$ and ${b = n}$ where ${n=|T_r|}$. In gradient based methods, the objective function ${E({\bf w})}$ under consideration is minimized by the iterative formula (\ref{eq:1}) where ${\it k}$ is the iteration count and ${\bf v}_{k+1}$ is the  update vector, which is defined for each gradient algorithm.
%\vspace{-1mm}
\begin{equation}\label{eq:1} 
{\bf w}_{k+1} = {\bf w}_k + {\bf v}_{k+1}.
\end{equation} 
In the following sections, we briefly discuss the full-batch BFGS quasi-Newton method and full-batch Nesterov's Accelerated quasi-Newton method in its full and limited memory forms. 
%We further extend to discuss relevant works on stochastic quasi-Newton methods.
We further extend to briefly discuss a stochastic BFGS method.
%\vspace{-4mm}
\iffalse
\begin{algorithm}[tb]
\caption{ BFGS Quasi-Newton Method}
\begin{algorithmic}[1]
\label{Algo:BFGS}
\REQUIRE  Terminate condition ${\varepsilon}$ and maximum iterations ${k_{max}}$
\ENSURE ${\bf w}_k $ to uniform random numbers and ${\bf H}_k$ to identity matrix.
\STATE $k \leftarrow 1$
%\STATE {\bf Initialize} ${\bf w}_1 $ to uniform random numbers and ${\bf H}_1$to identity matrix
\STATE Calculate $\nabla E({\bf w}_k)$
\WHILE{||$E({\bf w}_k)|| > {\bf \varepsilon}\;\;{\rm and}\;\; k < k_{max}$}
\iffalse
\STATE ${\bf {g}}_k\leftarrow-{\bf {\hat H}}_k \nabla E({\bf w}_k)$
\IF {$E({\bf w}_k + \mu {\bf v}_k+ {\bf {\hat g}}_k) \leq E({\bf w}_k+\mu {\bf v}_k) + \sigma \alpha_k \nabla E({\bf w}_k+\mu {\bf v}_k)^{\rm T} {\bf {\hat g}}_k$}
\STATE  $\alpha_k\leftarrow1$
\ELSE
\STATE $\alpha_k$ is computed by (34)			
\ENDIF
\fi
\STATE Determine $\alpha_k$ using Armijo line search
\STATE ${\bf v}_{k+1}\leftarrow -\alpha_k {\bf {H}}_k \nabla E({\bf w}_k)$\\
\STATE ${\bf w}_{k+1}\leftarrow{\bf w}_k + {\bf v}_{k+1}$\\
\STATE Calculate $\nabla E({\bf w}_{k+1})$\\
%\STATE ${\bf s}_{k}\leftarrow{\bf w}_{k+1} - ({\bf w}_k )$\\
%\STATE ${\bf y}_{k}\leftarrow \nabla E({\bf w}_{k+1})- \nabla E({\bf w}_k)$\\
\STATE Update ${\bf H}_{k+1}$ using (\ref{eq:5}) \\
\STATE $k \leftarrow k+1$
\ENDWHILE
\end{algorithmic}
\end{algorithm}
\fi

\hspace{-3mm}
\begin{minipage}{0.46\textwidth}
\centering
\begin{algorithm}[H]
\caption{ BFGS Method}
\begin{algorithmic}[1]
\label{Algo:BFGS}
\Require  ${\varepsilon}$ and ${k_{max}}$
\Ensure ${\bf w}_k \in \mathbb{R}^d$~and~${\bf H}_k$ = ${\bf I}$.
\State $k \leftarrow 1$
%\STATE {\bf Initialize} ${\bf w}_1 $ to uniform random numbers and ${\bf H}_1$to identity matrix
\State Calculate $\nabla E({\bf w}_k)$
\While {$||E({\bf w}_k)||>{\varepsilon}~{\rm and}~ k< k_{max}$}
\State ${\bf {g}}_k\leftarrow-{\bf {H}}_k \nabla E({\bf w}_k)$
\State Determine $\alpha_k$ by line search
\State ${\bf v}_{k+1}\leftarrow \alpha_k {\bf g}_k$
\State ${\bf w}_{k+1}\leftarrow{\bf w}_k + {\bf v}_{k+1}$
\State \underline{{Calculate $\nabla E({\bf w}_{k+1})$}}
%\vspace{1mm}
\State Update ${\bf H}_{k+1}$ using (\ref{eq:5})
\State $k \leftarrow k+1$
\EndWhile \underline{}

\end{algorithmic}
\end{algorithm}
\end{minipage}
\hspace{3mm}
\begin{minipage}{0.46\textwidth}
\begin{algorithm}[H]
    \centering
\caption{NAQ Method}
\begin{algorithmic}[1]
\label{Algo:NAQ}
\Require ${0<\mu<1}$, ${\varepsilon}$ and ${k_{max}}$
\Ensure ${\bf w}_k  \in \mathbb{R}^d$, ${\bf H}_k$ = ${\bf I}$ and ${\bf v}_k $ = ${0}$.
\State $k \leftarrow 1$
\While{$||E({\bf w}_k)|| > {\varepsilon}~{\rm and}~ k < k_{max}$}
\State  \underline{ {Calculate $\nabla E({\bf w}_k+\mu {\bf v}_k)$}}
\State ${\bf \hat {g}}_k\leftarrow-{\bf {\hat H}}_k \nabla E({\bf w}_k+\mu {\bf v}_k)$
\State Determine $\alpha_k$ by line search
\State ${\bf v}_{k+1}\leftarrow\mu {\bf v}_k +\alpha_k {\bf \hat{g}}_k$
\State ${\bf w}_{k+1}\leftarrow{\bf w}_k +{\bf v}_{k+1}$
\State \underline{ {Calculate $\nabla E({\bf w}_{k+1})$}}
\State Update ${\bf \hat H}_k$ using (\ref{eq:12}) 
\State $k \leftarrow k+1$
\EndWhile
\end{algorithmic}
\end{algorithm}
\end{minipage}

%\subsection{Stochastic quasi-Newton}
\subsection{BFGS quasi-Newton Method }%and LBFGS Method}
%In quasi-Newton (QN) methods, the Hessian matrix is computed by iterative approximations. The BFGS algorithm is one of the most popular quasi-Newton methods. Several improvements have been proposed to quasi-Newton methods that result in faster and better convergence.

Quasi-Newton methods utilize the gradient of the objective function to achieve superlinear or quadratic convergence. The Broyden-Fletcher-Goldfarb-Shanon (BFGS) algorithm is one of the most popular quasi-Newton methods for unconstrained optimization. The update vector of the quasi-Newton method is given as\begin{equation}
{\bf v}_{k+1} = \alpha_k {\bf g}_k,
\end{equation}
where ${\bf g}_k = -{\bf H}_k \nabla E({\bf w}_k)$ is the search direction. 
%\begin{equation}
%{\bf d}_k = -{\bf H}_k \nabla E({\bf w}_k).
%\end{equation}
The hessian matrix ${\bf H}_k$ is symmetric positive definite and is iteratively approximated by the following BFGS formula ~\cite{nocedal2006}.
\begin{equation}\label{eq:5}
{\bf H}_{k+1}= ( {\bf I}- {\bf s}_k {\bf y}_k^{\rm T}/{{\bf y}_k^{\rm T} {\bf s}_k}){\bf H}_k({\bf I}- {\bf y}_k {\bf s}_k^{\rm T}/{{\bf y}_k^{\rm T} {\bf s}_k})+ {\bf s}_k {\bf s}_k^{\rm T}/{{\bf y}_k^{\rm T} {\bf s}_k},
\end{equation}
where ${\bf I}$ denotes identity matrix,
\begin{equation}
{\bf s}_k = {\bf w}_{k+1} - {\bf w}_k ~~{\rm and}~~ {\bf y}_ k = \nabla E ( {\bf w}_{k+1} ) - \nabla E ({\bf w}_k).
\end{equation}
%\begin{equation}
%{\bf y}_k = \nabla E ( {\bf w}_{k+1} ) - \nabla E ({\bf w}_k), 
%\end{equation}
%\begin{equation}
%\rho_k = \frac{1}{{\bf y}_k^{\rm T} {\bf s}_k}.
%\end{equation}
The BFGS quasi-Newton algorithm is shown in Algorithm 1.%\ref{Algo:BFGS}

\iffalse
\begin{algorithm}[tb]
\caption{ Nesterov's Accelerated Quasi-Newton (NAQ)}
\begin{algorithmic}[1]
\label{Algo:NAQ}
\Require Momemtum parameter ${0<\mu<1}$, terminate condition ${\varepsilon}$ and maximum iterations ${k_{max}}$
\Ensure ${\bf w}_k $ to uniform random numbers, ${\bf H}_k$ to identity matrix and ${\bf v}_k $ to zero vector.
\State $k \leftarrow 1$
%\STATE {\bf Initialize} ${\bf w}_1 $ to uniform random numbers and ${\bf H}_1$to identity matrix
\While{$||E({\bf w}_k)|| > {\bf \varepsilon}\;\;{\rm and}\;\; k < k_{max}$}
\State Calculate $\nabla E({\bf w}_k+\mu {\bf v}_k)$
\State ${\bf {g}}_k\leftarrow-{\bf {\hat H}}_k \nabla E({\bf w}_k+\mu {\bf v}_k)$
\State Determine $\alpha_k$ using line search
\State ${\bf v}_{k+1}\leftarrow{\bf w}_k +\mu {\bf v}_k +\alpha_k {\bf {g}}_k$
\State ${\bf w}_{k+1}\leftarrow{\bf w}_k +{\bf v}_{k+1}$
\State Calculate $\nabla E({\bf w}_{k+1})$
%\STATE ${\bf p}_{k}\leftarrow{\bf w}_{k+1} - ({\bf w}_k +\mu{\bf v}_{k})$\\
%\STATE ${\bf q}_{k}\leftarrow \nabla E({\bf w}_{k+1})- \nabla E({\bf w}_k+\mu {\bf v}_k)$
\State Update ${\bf \hat H}_k$ using (\ref{eq:12})
\State $k \leftarrow k+1$
\EndWhile
\end{algorithmic}
\end{algorithm}
\fi

\subsubsection{Limited Memory BFGS (LBFGS): }
LBFGS is a variant of the BFGS quasi-Newton method, designed for solving large-scale optimization problems. As the scale of the neural network model increases, the {\it O${(d^2)}$} cost of storing and updating the Hessian matrix ${\bf H}_k$ is expensive \cite{schraudolph2007stochastic}. In the limited memory version, the Hessian matrix is defined by applying {\it m} BFGS updates using only the last {\it m} curvature pairs ${\{{\bf s}_k,{\bf y}_k\}}$. %Thus the limited memory scheme - (o)LBFGS proves to be effective as the inverse Hessian is based on only the last m stored vectors. 
As a result, the computational cost is significantly reduced and the storage cost is down to {\it O${(md)}$} where ${d}$ is the number of parameters and ${m}$ is the memory size.%which is drastically reduced especially when ${m\ll n}$ 

\subsection{Nesterov's Accelerated Quasi-Newton Method}

Several modifications have been proposed to the quasi-Newton method to obtain stronger convergence. The Nesterov's Accelerated Quasi-Newton (NAQ) ~\cite{ninomiya2017novel} method achieves faster convergence compared to the standard quasi-Newton methods by quadratic approximation of the objective function at ${\bf w}_k+\mu {\bf v}_k$ and by incorporating the Nesterov's accelerated gradient $\nabla E({\bf w}_k+\mu {\bf v}_k)$ in its Hessian update. The derivation of NAQ is briefly discussed as follows.

 %\subsection{Standard NAQ}
 %In this section we briefly discuss the derivation of NAQ.
 Let $\Delta {\bf w}$ be the vector $\Delta {\bf w}={\bf w}-\left({\bf w}_k+\mu{\bf v}_k \right)$. The quadratic approximation of the objective function at ${\bf w}_k+\mu{\bf v}_k$ is defined as,
\begin{equation}\label{eq:9}
%\begin{split}
%E\left({\bf w}\right) \simeq E\left({\bf w}_k+\mu_k{\bf v}_k \right)+ \nabla E\left({\bf w}_k+\mu_k{\bf v}_k \right)^{\rm T} \Delta{\bf w}\\+\frac{1}{2} \Delta {\bf w}^{\rm T}
E({\bf w}) \simeq E({\bf w}_k+\mu{\bf v}_k)+ \nabla E({\bf w}_k+\mu{\bf v}_k)^{\rm T} \Delta{\bf w}\\+\frac{1}{2} \Delta {\bf w}^{\rm T} \nabla^2 E({\bf w}_k+\mu{\bf v}_k) \Delta {\bf w}.
%\end{split}
\end{equation}
The minimizer of this quadratic function is explicitly given by
\begin{equation}\label{eq:10}
%\begin{split}
\Delta {\bf w}=-\nabla^2 E\left({\bf w}_k+\mu{\bf v}_k \right)^{-1} \nabla E\left({\bf w}_k+\mu{\bf v}_k \right).
%\end{split}
\end{equation}
Therefore the new iterate is defined as
\begin{equation}\label{eq:11}
%\begin{split}
{\bf w}_{k+1}=\left({\bf w}_k+\mu{\bf v}_k \right)
-\nabla^2 E\left({\bf w}_k+\mu{\bf v}_k \right)^{-1} \nabla E\left({\bf w}_k+\mu{\bf v}_k \right).
%\end{split}
\end{equation}
This iteration is considered as Newton method with the momentum term $\mu{\bf v}_k$. The inverse of Hessian $\nabla^2 E({\bf w}_k + \mu {\bf v}_k)$ is approximated by the matrix ${\bf \hat H}_{k+1}$ using the update equation (\ref{eq:12})

\begin{equation}\label{eq:12}
{\bf {\hat H}}_{k+1}= ( {\bf I}-{\bf p}_k {\bf q}_k^{\rm T}/{{\bf q}_k^{\rm T} {\bf p}_k}){\bf {\hat H}}_k({\bf I}- {\bf q}_k {\bf p}_k^{\rm T}/{{\bf q}_k^{\rm T} {\bf p}_k})+ {\bf p}_k {\bf p}_k^{\rm T}/{{\bf q}_k^{\rm T} {\bf p}_k},
\end{equation}
 where \begin{equation}\label{eq:13}
{\bf p}_k = {\bf w}_{k+1} - ({\bf w}_k+ \mu{\bf v}_k) ~~{\rm and}~~{\bf q}_k = \nabla E ( {\bf w}_{k+1} ) - \nabla E ({\bf w}_k+ \mu {\bf v}_k).
\end{equation}
(\ref{eq:12}) is derived from the secant condition ${\bf q}_k=({\bf \hat H}_{k+1})^{-1}{\bf p}_k$ and the rank-2 updating formula ~\cite{ninomiya2017novel}.
%\begin{equation}\label{eq:16}
%{\bf q}_k=({\bf \hat H}_{k+1})^{-1}{\bf p}_k.
%\end{equation}
 It is proved that the Hessian matrix ${\bf \hat H}_{k+1}$ updated by  (\ref{eq:12}) is a positive definite symmetric matrix given ${\bf \hat H}_k$ is initialized to identity matrix~\cite{ninomiya2017novel}. Therefore, the update vector of NAQ can be written as:
\begin{equation}\label{eq:17}
{\bf v}_{k+1} = \mu {\bf v}_k + \alpha_k {\bf \hat{g}}_k,
\end{equation} 
%\begin{equation}\label{eq:18}
% {\bf { g}}_k=-{\bf {\hat H}}_k \nabla E({\bf w}_k+\mu{\bf v}_k).
%\end{equation} 
where ${\bf \hat{ g}}_k=-{\bf {\hat H}}_k \nabla E({\bf w}_k+\mu{\bf v}_k)$ is the search direction. The NAQ algorithm is given in Algorithm 2. %\ref{Algo:NAQ}. 
Note that the gradient is computed twice in one iteration. This increases the computational cost compared to the BFGS quasi-Newton method. However, due to acceleration by the momentum and Nesterov's gradient term, NAQ is faster in convergence compared to BFGS.

 \begin{algorithm}[tb]
\caption{ Direction Update}
\begin{algorithmic}[1]
\label{Algo:dirUp}
\Require current gradient ${\nabla E({\bf \theta}_k)}$, memory size ${m}$, curvature pair (${\sigma_{k-i}}$, ${\gamma_{k-i}}$)  $\forall { i=1, 2, ..., min(k-1,m) }$  where ${\sigma_{k}}$ is the difference of current and previous weight vector  and ${\gamma_{k}}$ is the difference of current and previous gradient vector 

\State ${\bf {\eta}}_k=-\nabla E({\bf \theta}_k)$
\For {$i:=1, 2, ..., {\rm min}(m,k-1)$}
\State ${ \beta}_i=({{\bf \sigma}_{k-i}^{\rm T} {\bf {\eta}}_k})/({{\bf \sigma}_{k-i}^{\rm T} {\bf \gamma}_{k-i}})$
\State ${\bf {\eta}}_k={\bf {\eta}}_k-{ \beta}_i {\bf \gamma}_{k-i}$
\EndFor
\If {$ k>1$}
\State ${\bf {\eta}}_k={\bf {\eta}}_k({{\bf \sigma}_k^{\rm T} {\bf \gamma}_k}/{{\bf \gamma}_k^{\rm T} {\bf \gamma}_k})$
\EndIf
\For {$i:k-{\rm min}(m,(k-1)),\ldots,k-1,k$}
\State${\tau}=({{\bf \gamma}_i^{\rm T} {\bf {\eta}}_k})/({{\bf \gamma}_i^{\rm T} {\bf \sigma}_i})$
\State${\bf {\eta}}_k={\bf {\eta}}_k-({\beta}_i-{\tau}){\bf \sigma}_i$
\EndFor
\State{\bf return}\;${\bf {\eta}}_k$
\end{algorithmic}
\end{algorithm}

%\vspace{-2mm}
\subsubsection{Limited Memory NAQ (LNAQ)} Similar to LBFGS method, LNAQ~\cite{LNAQ_shah} is the limited memory variant of NAQ that uses the last {\it m} curvature pairs ${\{ {{\bf p}_k,{\bf q}_k}\}}$. In the limited-memory form note that the curvature pairs that are used incorporate the momemtum and Nesterov's accelerated gradient term, thus accelerating LBFGS. Implementation of LNAQ algorithm can be realized by omitting steps 4 and 9 of Algorithm 2 %\ref{Algo:NAQ}
 and determining the search direction ${\bf \hat g}_k$ using the two-loop recursion~\cite{nocedal2006} shown in Algorithm 3. % \ref{Algo:dirUp}.  
 The last {\it m} vectors of ${\bf p}_k$ and ${\bf q}_k$ are stored and used in the direction update.

\subsection{Stochastic BFGS quasi-Newton Method (oBFGS)}%and LBFGS Method}
The online BFGS method proposed by Schraudolph et al in \cite{schraudolph2007stochastic} is a fast and scalable stochastic quasi-Newton method suitable for convex functions. The changes proposed to the BFGS method in  \cite{schraudolph2007stochastic} to work well in a stochastic setting are discussed as follows. The line search is replaced with a gain schedule such as  
\begin{equation}\label{eq:gain}
{\alpha_k} = {\tau}/({\tau+k})\cdot\alpha_0, 
\end{equation}
where $\alpha_0, \tau>0 $ provided the Hessian matrix is positive definite%${\bf H}_k$${ > 0}$
, thus restricting to convex optimization problems. Since line search is eliminated, the first parameter update is scaled by a small value. Further, to improve the performance of oBFGS, the step size is divided by an analytically determined constant ${c}$. An important modification is the computation of ${\bf y}_k$, the difference of the last two gradients is computed on the same sub-sample ${X_k}$\cite{schraudolph2007stochastic,mokhtari2014res} as given below,
\begin{equation}\label{eq:yk}
{\bf y}_k =  \nabla E ( {\bf w}_{k+1}, X_k ) - \nabla E ({\bf w}_k, X_k).
\end{equation}
 This however doubles the cost of gradient computation per iteration but is shown to outperform natural gradient descent for all batch sizes \cite{schraudolph2007stochastic}. The oBFGS algorithm is shown in Algorithm 4. In this paper, we introduce direction normalization as shown in step 5, details of which are discussed in the next section.  %\ref{Algo:oBFGS}.

\subsubsection{Stochastic Limited Memory BFGS (oLBFGS)}  \cite{schraudolph2007stochastic} further extends the oBFGS method to limited memory form by determining the search direction ${\bf g}_k$ using the two-loop recursion (Algorithm 3% \ref{Algo:dirUp}
). The Hessian update is omitted and instead the last ${m}$ curvature pairs ${\bf s}_k$ and ${\bf y}_k$ are stored. This brings down the computation complexity to ${2bd+6md}$ where ${b}$ is the batch size, ${d}$ is the number of parameters, and ${m}$ is the memory size. To improve the performance by averaging sampling noise step 7 of Algorithm 3 %\ref{Algo:dirUp} 
is replaced by (\ref{eq:oLBFGS}) where ${\bf \sigma}_k$ is ${\bf s}_k$ and ${\bf \gamma}_k$ is ${\bf y}_k$.

\begin{equation}
\label{eq:oLBFGS}
{\bf \eta}_k = 
\begin{cases}
~~~~~~~~~~~~~~~~~~~~~~~~~~~~~~~~~~~\epsilon {\bf \eta}_k ~~~~~~~~~~~{\rm if~ } {\it k}=1,\\
\displaystyle\frac{{\bf \eta}_k}{\rm min(k,m)}\displaystyle\sum^{\rm min(k,m)}_{i=1} \frac{{\bf \sigma}_{k-i}^{\rm T} {\bf \gamma}_{k-i}}{{\bf \gamma}_{k-i}^{\rm T} {\bf \gamma}_{k-i}}~~~~~~~~~{\rm otherwise}.
\end{cases}
\end{equation} 

\section{Proposed Algorithm - oNAQ and oLNAQ}

%The stochastic quasi-Newton method proposed in ~\cite{schraudolph2007stochastic} describes a fast and scalable method that systematically modifies the BFGS quasi-Newton method in both its full and limited memory forms. 
The oBFGS method proposed in ~\cite{schraudolph2007stochastic} computes the gradient of a sub-sample minibatch ${X_k}$ twice in one iteration. This is comparable with the inherent nature of NAQ which also computes the gradient twice in one iteration.  Thus by applying suitable modifications to the original NAQ algorithm, we achieve a stochastic version of the Nesterov's Accelerated Quasi-Newton method. 
The proposed modifications for a stochastic NAQ method is discussed below in its full and limited memory forms. 
%\indent In this paper, we focus our comparison of the proposed method with the stochastic BFGS method in ~\cite{schraudolph2007stochastic}

%The stochastic quasi-Newton method described in ~\cite{schraudolph2007stochastic} computes the gradient of a sub-sample minibatch ${X_k}$ twice in one iteration. This is comparable with inherent nature of NAQ which also computes the gradient twice in one iteration.  Thus by applying suitable modifications, it is possible to achieve a stochastic version of the Nesterov's Accelerated Quasi-Newton method. 

\subsection{Stochastic NAQ (oNAQ)}
The NAQ algorithm computes two gradients, $\nabla E({\bf w}_k+\mu{\bf v}_k)$ and $\nabla E({\bf w}_{k+1}) $ to calculate ${\bf q}_k$ as shown in (\ref{eq:13}). On the other hand, the oBFGS method proposed in ~\cite{schraudolph2007stochastic} computes the gradient $\nabla E({\bf w}_k, X_k)$ and $\nabla E({\bf w}_{k+1},X_k) $ to calculate ${\bf y}_k$ as shown in (\ref{eq:yk}). Therefore, oNAQ can be realised by changing steps 3 and 8 of Algorithm 2 to calculate $\nabla E({\bf w}_k+\mu{\bf v}_k, X_k)$ and $\nabla E({\bf w}_{k+1}, X_k) $. Thus in oNAQ, the ${\bf q}_k$ vector is given by (\ref{new:qk}) where $\lambda {\bf p}_k$ is used to guarantee numerical stability \cite{zhang2005globally,dai2002convergence,indrapriyadarsini2018implementation}.   \begin{equation}{\bf q}_k = \label{new:qk}{\nabla E({\bf w}_{k+1}, X_k) - \nabla E({\bf w}_k+\mu{\bf v}_k, X_k) + \lambda {\bf p}_k },\end{equation} 

Further, unlike in full batch methods, the updates in stochastic methods have high variance resulting in the objective function to fluctuate heavily. This is due to the updates being performed based on small sub-samples of data. This can be seen more prominently in case of the limited memory version where the updates are based only on ${m}$ recent curvature pairs. Thus in order to improve the stability of the algorithm, we introduce direction normalization as   \begin{equation}\label{eq:dirNorm}
{\bf \hat {g}}_k  = {{\bf \hat {g}}_k }/{||{\bf \hat {g}}_k||_{2}}, \end{equation}
where ${||{\bf \hat {g}}_k||_{2}}$ is the ${l_2}$ norm of the search direction ${{\bf \hat{g}}_k}$. Normalizing the search direction at each iteration ensures that the algorithm does not move too far away from the current objective~\cite{li2018implementation}. Fig.\ref{fig:dirNorm} illustrates the effect of direction normalization on oBFGS and the proposed oNAQ method. %Note that in all the figures, 
The solid lines indicate the moving average.  As seen from the figure, direction normalization improves the performance of both oBFGS and oNAQ. Therefore, in this paper we include direction normalization for oBFGS also. % on the standard MNIST dataset with a NN structure 784-30-20-10.

The next proposed modification is with respect to the step size. In full batch methods, the step size or the learning rate is usually determined by line search methods satisfying either Armijo or Wolfe conditions. However, in stochastic methods, line searches are not quite effective since search conditions apply global validity. This cannot be assumed when using small local sub-samples\cite{schraudolph2007stochastic}. Several studies show that line search methods does not necessarily ensure global convergence and have proposed methods that eliminate line search~\cite{zhang2005globally,dai2002convergence,indrapriyadarsini2018implementation}. Moreover, determining step size using line search methods involves additional function computations until the search conditions such as the Armijo or Wolfe condition is satisfied. Hence we determine the step size using a simple learning rate schedule. Common learning rate schedules are polynomial decays and exponential decay functions. In this paper, we determine the step size using a polynomial decay schedule\cite{zinkevich2003online}   {\begin{equation}\label{eq:alpha}
 \alpha_k= {\alpha_0}/{\sqrt[]{\mathstrut {k}}},
\end{equation}
where ${\alpha_0}$ is usually set to 1. If the step size is too large, which is the case in the initial iterations, the learning can become unstable. This is stabilized by direction normalization. A comparison of common learning rate schedules are illustrated in Fig. \ref{fig:alpha}

The proposed stochastic NAQ algorithm is shown in Algorithm 5. % \ref{Algo:oNAQ}.
Note that the gradient is computed twice in one iteration, %however on the same sub-sample ${X_k}$. The
thus making the computational cost same as that of the stochastic BFGS (oBFGS) proposed in \cite{schraudolph2007stochastic}.   %But Nesterov's acceleration and direction normalization of the proposed algorithm results in faster convergence compared to existing algorithms.  
%The gradient is computed twice in one iteration, however on the same sub-sample ${X_k}$. We introduce direction normalization in step 4 in order to improve the stability of the algorithm.  Also, the step size ${\alpha_k}$ is determined using a simple learning rate schedule as discussed below.

%${\alpha_0}$ acts is a scaling factor to avoid the step size from diminishing to a very small value as the number of iterations increase. 
%\vspace{-3mm}
\hspace{-4mm}
\begin{minipage}{0.46\textwidth}
\centering
\begin{algorithm}[H]
\caption{ oBFGS Method}
\begin{algorithmic}[1]
\label{Algo:oBFGS}
\Require minibatch ${X_k}$,  ${k_{max}}$ and $ {\lambda \geq 0}$, 
\Ensure $ {\bf w}_k  \in  \mathbb{R}^d$, ${{\bf { H}}_k = \epsilon  \bf I}$ and ${{\bf v}_k = 0}$ 
\State $k \leftarrow 1$
\While{$ k < k_{max}$}
\State ${\bf\nabla E}_1 \leftarrow \nabla E({\bf w}_k, X_k)$
\State ${\bf {g}}_k\leftarrow-{\bf { H}}_k \nabla E({\bf w}_k,X_{k})$
\State ${\bf {g}}_k  = {{\bf {g}}_k }/{||{\bf {g}}_k||_{2}}$ 
\State Determine $\alpha_k$ using (\ref{eq:gain})
\State ${\bf v}_{k+1}\leftarrow\alpha_k {\bf {g}}_k$
\State ${\bf w}_{k+1}\leftarrow{\bf w}_k +{\bf v}_{k+1}$
\State ${\bf\nabla E}_2 \leftarrow\nabla E({\bf w}_{k+1}, X_k)$
\State ${\bf s}_{k}\leftarrow{\bf w}_{k+1} - {\bf w}_k $
%\State ${\bf y}_{k}\leftarrow {\nabla E({\bf w}_{k+1}, X_{k})} -{\nabla E({\bf w}_k , X_{k})} + \lambda {\bf s}_{k}$
\State ${\bf y}_{k} ~\leftarrow {\bf\nabla E}_2 - {\bf\nabla E}_1 + \lambda {\bf s}_{k}$
\State Update ${\bf H}_k$ using (\ref{eq:5}) 
\State $k \leftarrow k+1$
\EndWhile

\end{algorithmic}
\end{algorithm}
\end{minipage}
\hspace{3mm}
\begin{minipage}{0.46\textwidth}
\begin{algorithm}[H]
\caption{ Proposed oNAQ Method}
\begin{algorithmic}[1]
\label{Algo:oNAQ}
\Require minibatch ${X_k}$, ${0<\mu<1}$ and ${k_{max}}$
\Ensure $ {\bf w}_k  \in  \mathbb{R}^d$, ${{\bf {\hat H}}_k = \epsilon  \bf I}$ and ${{\bf v}_k = 0}$ 
\State $k \leftarrow 1$
\While{$ k < k_{max}$}
\State ${\bf \nabla E}_1  \leftarrow \nabla E({\bf w}_k+\mu {\bf v}_k, X_k)$
\State ${\bf \hat{g}}_k\leftarrow-{\bf {\hat H}}_k \nabla E({\bf w}_k+\mu {\bf v}_k, X_{k})$
%\State Direction normalization as in (\ref{eq:dirNorm})
\State ${\bf \hat {g}}_k  = {{\bf \hat {g}}_k }/{||{\bf \hat {g}}_k||_{2}}$ 
\State Determine $\alpha_k$ using (\ref{eq:alpha})
\State ${\bf v}_{k+1}\leftarrow\mu {\bf v}_k +\alpha_k {\bf \hat{g}}_k$
\State ${\bf w}_{k+1}\leftarrow{\bf w}_k +{\bf v}_{k+1}$
\State ${\bf \nabla E}_2 \leftarrow \nabla E({\bf w}_{k+1}, X_k)$
\State ${\bf p}_{k}\leftarrow{\bf w}_{k+1} -({\bf w}_k + \mu {\bf v}_{k})$
%\State ${\bf q}_{k}\leftarrow {\nabla E({\bf w}_{k+1}, X_{k})}-{\nabla E({\bf w}_k+\mu {\bf v}_{k}, X_{k})}+\lambda {\bf p}_{k}$
\State ${\bf q}_{k}\leftarrow {\bf \nabla E}_2 - {\bf \nabla E}_1 +\lambda {\bf p}_{k}$
\State Update ${\bf \hat H}_k$ using (\ref{eq:12}) 

\State $k \leftarrow k+1$
\EndWhile
\end{algorithmic}
\end{algorithm}
\end{minipage}

%\vspace{-5mm}
\begin{figure}
\centering
\begin{minipage}{.5\textwidth}
  \centering
\captionsetup{justification=centering}
  \captionsetup{width=0.95\linewidth}
  \includegraphics[width=1\linewidth]{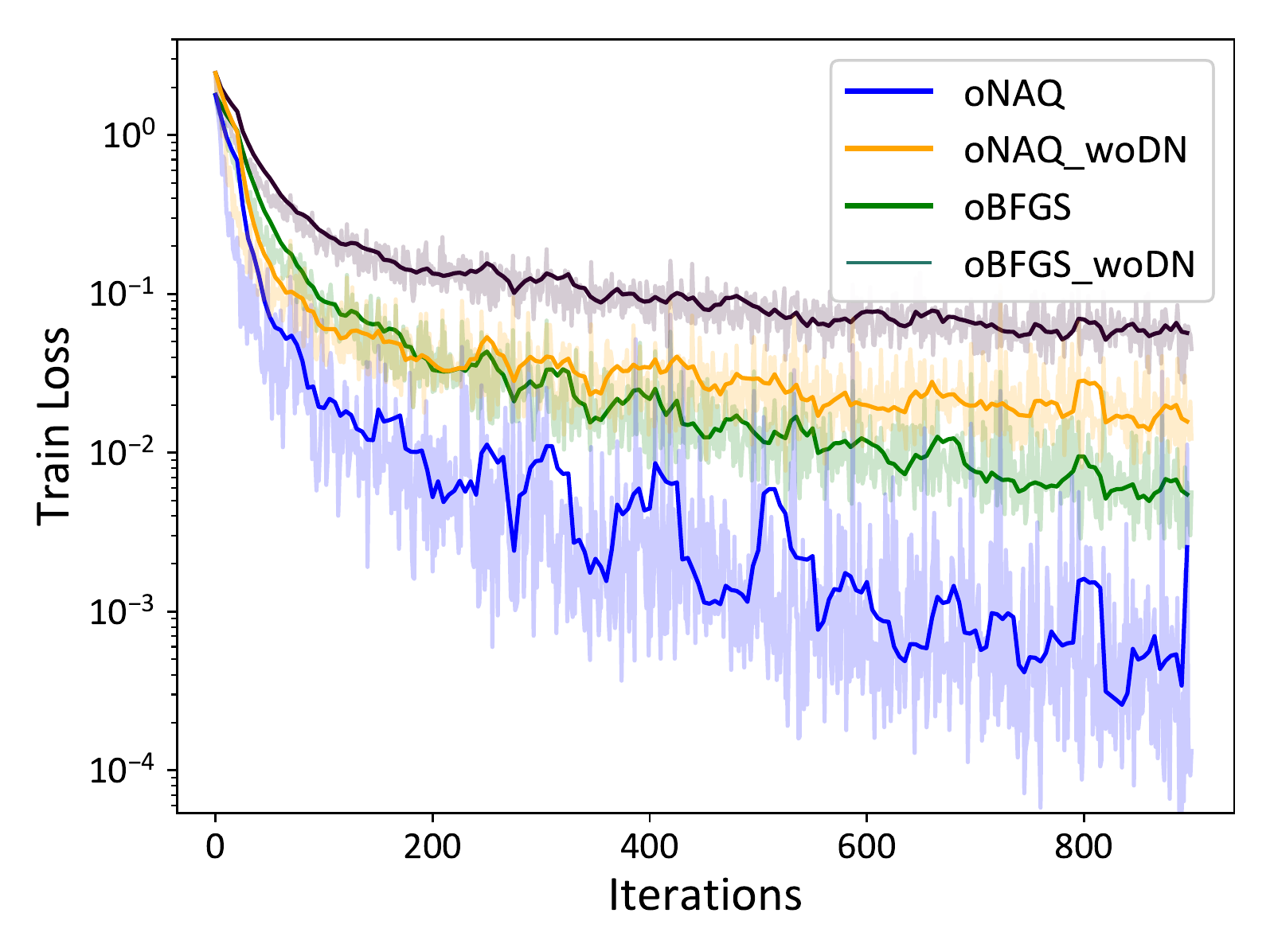}
  \captionof{figure}{Effect of direction normalization on 8x8 MNIST with b = 64 and ${\rm \mu=0.8}$. }
  \label{fig:dirNorm}
\end{minipage}%
\begin{minipage}{0.5\textwidth}
  \centering
\captionsetup{justification=centering}
  \captionsetup{width=0.9\linewidth}
  \hspace{-3mm}
  \includegraphics[width=1\linewidth]{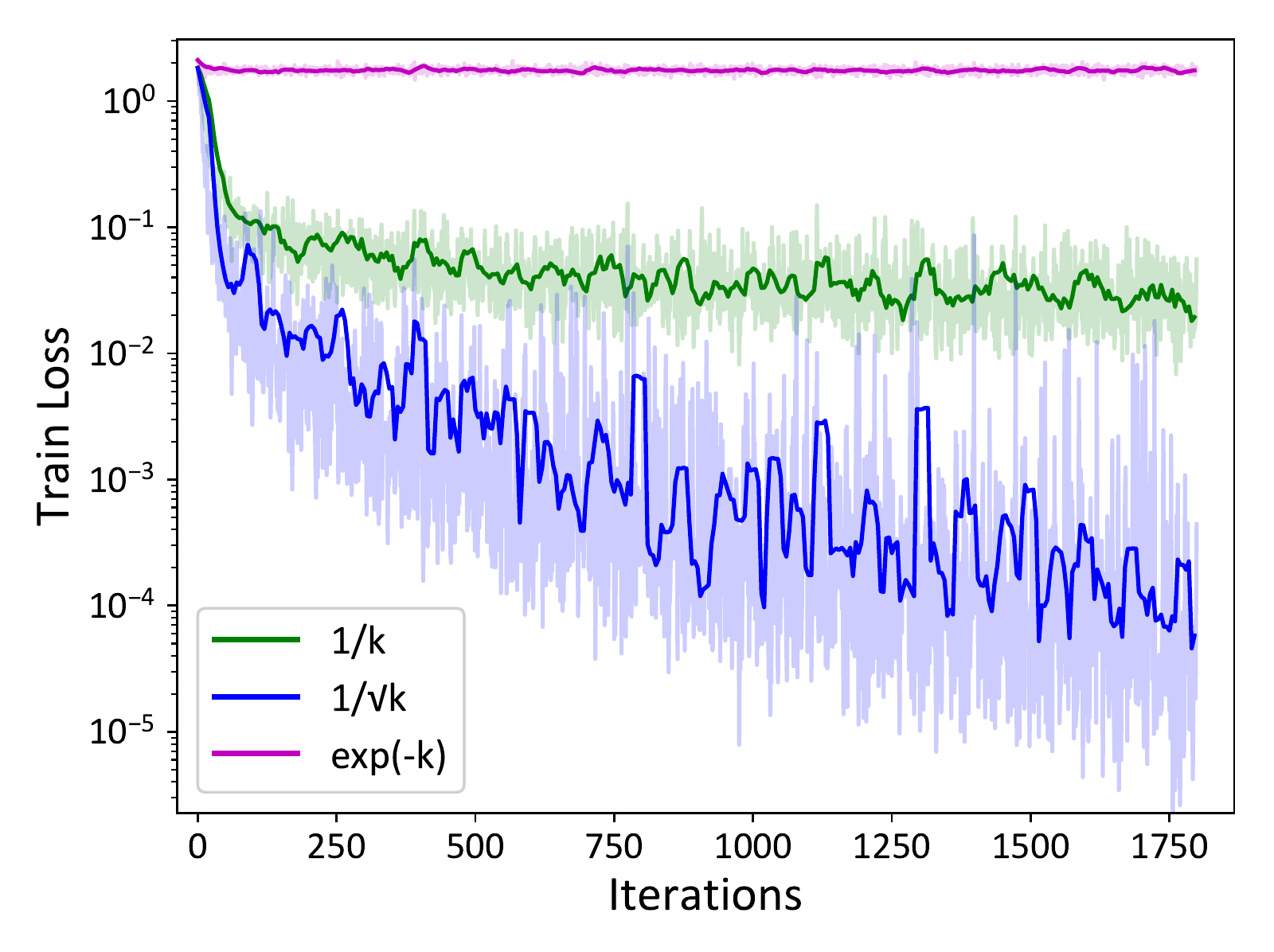}
  \captionof{figure}{Comparison of ${\alpha_k}$ schedules on 8x8 MNIST with b = 64 and ${\rm \mu=0.8}$.  }
  \label{fig:alpha}
\end{minipage}
\end{figure}
\subsection{Stochastic Limited-Memory NAQ (oLNAQ)}
Stochastic LNAQ can be realized by making modifications to Algorithm 5 %\ref{Algo:oNAQ} 
similar to LNAQ. The search direction ${\bf \hat g}_k$ in step 4 is determined by Algorithm 3. %\ref{Algo:dirUp}. 
oLNAQ like LNAQ uses the last ${m}$ curvature pairs ${ \{{\bf p}_k,{\bf q}_k\}}$ to estimate the Hessian matrix instead of storing and computing on a ${d}$x${d}$ matrix. Therefore, the implementation of oLNAQ does not require initializing or updating the Hessian matrix. Hence step 12 of Algorithm 5 %\ref{Algo:oNAQ} 
is replaced by storing the last ${m}$ curvature pairs ${ \{{\bf p}_k,{\bf q}_k\}}$. Finally, in order to average out the sampling noise in the last ${m}$ steps, we replace step 7 of Algorithm 3 %\ref{Algo:dirUp} 
by eq. (\ref{eq:oLBFGS}) 
%(\ref{eq:oLNAQ}) 
where ${\bf \sigma}_k$ is ${\bf p}_k$ and ${\bf \gamma}_k$ is ${\bf q}_k$.
%In addtion, step 10 of Algorithm 2 is replaced by (\ref{eq:oLNAQ}) to average out the sampling noise in the last {\it m} steps. 
Note that an additional ${2md}$ evaluations are required to compute (\ref{eq:oLBFGS}). However the overall computation cost of oLNAQ is much lesser than that of oNAQ and the same as oLBFGS.

%\vspace{-8mm}
\section{Simulation Results}
We illustrate the performance of the proposed stochastic methods oNAQ and oLNAQ on four benchmark datasets - two classification and two regression problems. For the classification problem we use the 8x8 MNIST and 28x28 MNIST datasets and for the regression problem we use the Wine Quality \cite{cortez2009modeling} and CASP \cite{rana2013physicochemical} datasets. We evaluate the performance of the classification tasks on a multi-layer neural network (MLNN) and a simple convolution neural network (CNN). The algorithms oNAQ, oBFGS, oLNAQ and oLBFGS are implemented in Tensorflow using the ScipyOptimizerInterface class.  Details of the simulation are given in Table \ref{tab2}. 
%For classification problem we use the 8x8 and 28x28 MNIST dataset and for th 
%We consider a simple multilayer neural network with 2 hidden layers. 
%\vspace{-3mm}
\subsection{Multi-Layer Neural Networks - Classification Problem}
We evaluate the performance of the proposed algorithms for classification of handwritten digits using the 8x8 MNIST \cite{alpaydin1998optical} and 28x28 MNIST dataset \cite{mnist2010data}. We consider a simple MLNN with two hidden layers. ReLU activation function and softmax cross-entropy loss function is used. Each layer except the output layer is batch normalized. %The simulations are run on a Intel(R) Core(TM) i5-8250U CPU @ 1.60 GHz 8GB memory. 
%Intel(R) Core(TM) i7-3770 CPU @ 3.40 GHz 12GB memory.
%Intel(R) Core(TM) i7-5820K CPU @ 3.30 GHz 16GB memory. 
%Due to system limitations, 
%Due to large number of parameters, we illustrate the performace of the full memory oNAQ on the smaller 8x8 MNIST dataset  and the limited memory oLNAQ on the larger 28x28 MNIST dataset.  %The step size scaling factor ${\alpha_0}$ is usually set to 1. %However, it is observed that as the number of epochs ({\it ep}) increases, the step size ${\alpha_k}$ can become exceedingly small. In order to resolve this issue, ${\alpha_k}$ is scaled by the epoch count by setting ${\alpha_0}$ to {\it ep}
%In order to have a fair comparison, the results of the proposed algorithm are compared with oBFGS and oLBFGS with direction normalization.
 
%\subsection{Experiment: Multi Layer Neural Networks}
\begin{table}
\begin{center}
\caption{Details of the Simulation - MLNN.}\label{tab2}
\begin{tabular}{c c c c c c }
\hline
\\[-3mm]
{\;\;\;  \;\;\;} &  {\;\;\;  8x8 MNIST \;\;\;} & {\;\;\; 28x28 MNIST \;\;\;}& {\;\;\;\; Wine Quality \;\;\;\;} & {\;\;\;\; CASP \;\;\;\;}\\
\hline
\\[-3mm]
{task } & classification & classification & regression & regression \\
{input } & 8x8 & 28x28 & 11 & 9\\
%{\bf NN structure} & 64-20-10-10 & 784-30-20-10 \\
{MLNN structure} & 64-20-10-10 & 784-100-50-10 & 11-10-4-1 & 9-10-6-1\\
{parameters (\it d)} &  1,620 & 84,060 & 169 & 173\\
%{\bf d} &  1620 & 24380\\
%{\bf d} &  1,620 & 936,330\\
{train set} &  1,198 & 55,000 & 3,918 & 36,584\\
{test set} & 599 & 10,000 & 980 & 9,146\\
{classes/output } & 10 & 10 & 1 & 1 \\
{momentum ($\mu$)} & 0.8 & 0.85 & 0.95 & 0.95\\
{batch size ($b$)} & 64 & 64/128 & 32/64 & 64/128\\
{memory ($m$)} & 4 & 4 & 4 & 4\\
\hline
\end{tabular}
\end{center}
%\vspace{-5mm}
\end{table}

\begin{figure}[h!]
  \centering
  \hspace{-3mm}
  \begin{subfigure}[b]{0.5\linewidth}
    \includegraphics[width=\linewidth]{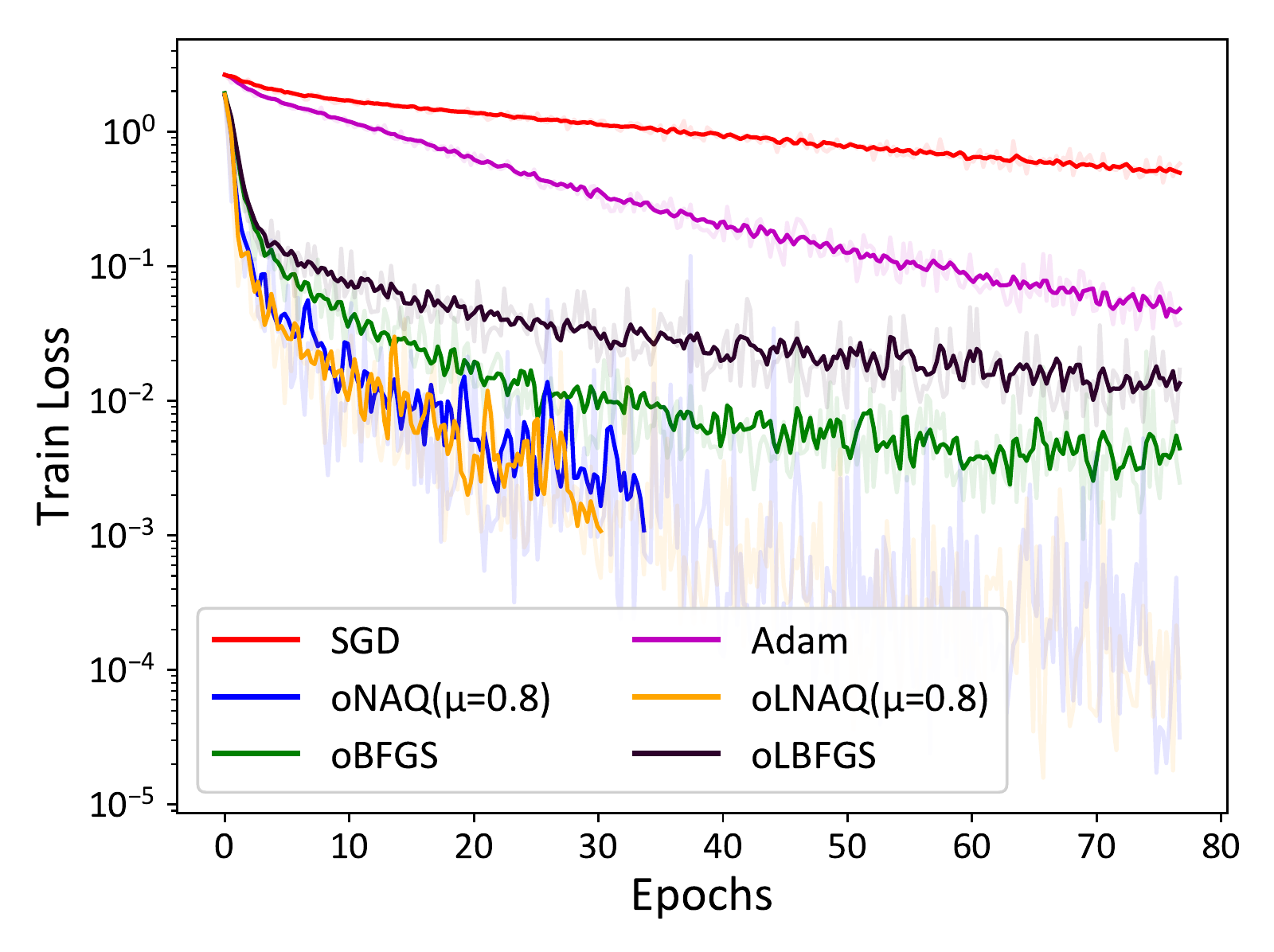}
    %\caption{}
  \end{subfigure}
  \begin{subfigure}[b]{0.5\linewidth}
    \includegraphics[width=\linewidth]{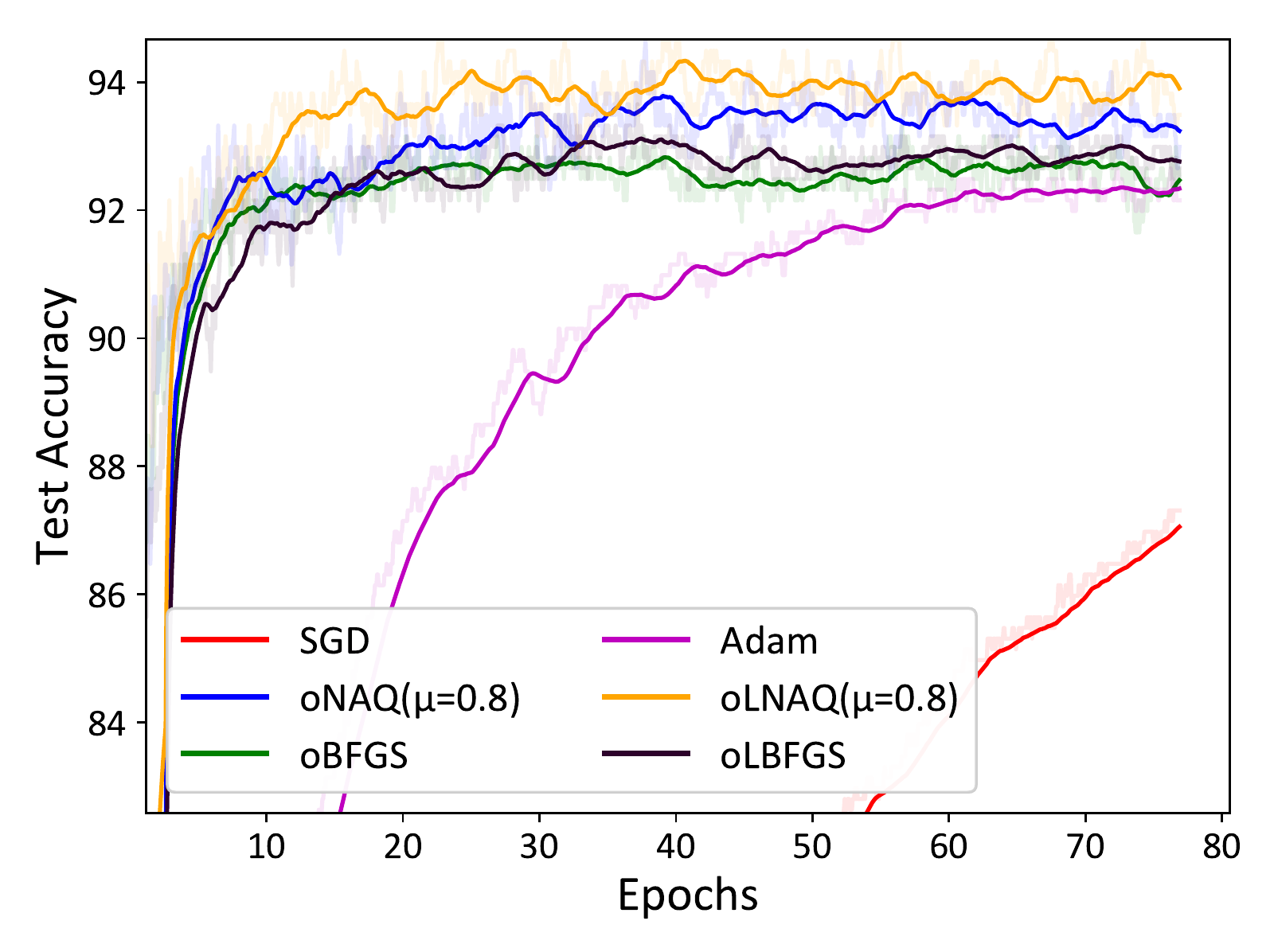}
    %\caption{}
  \end{subfigure}
  \caption{Comparision of train loss and test accuracy versus number of epochs required for convergence of 8x8 MNIST data with a maximum of 80 epochs. }
  \label{fig:eps}
\end{figure}

%\vspace{-3mm}
\subsubsection{Results on 8x8 MNIST Dataset}
We evaluate the performance of oNAQ and oLNAQ on a reduced version of the MNIST dataset in which each sample is an 8x8 image representing a handwritten digit \cite{alpaydin1998optical}.
Fig. \ref{fig:eps} shows the number of epochs required to converge to a train loss of ${<10^{-3}}$ and its corresponding test accuracy for a batch size ${b=64}$. The maximum number of epochs is set to 80. % for a batch size ${b=64}$, $\mu=0.8$ and limited memory $m=4$.
As seen from the figure, it is clear that oNAQ and oLNAQ require fewer epochs compared to oBFGS, oLBFGS, Adam and SGD. In terms of compuation time, o(L)BFGS and o(L)NAQ require longer time compared to the first order methods. This is  due to the Hessian computation and twice gradient calculation. Further,  the oBFGS and oNAQ per iteration time difference compared to first order methods is much larger than that of the limited memory algorithms with memory $m = 4$.  This can be seen from Fig. \ref{fig:oNAQ} which shows the comparison of train loss and test accuracy versus time for 80 epochs. It can be observed that for the same time, the second order methods perform significantly better compared to the first order methods, thus confirming that the extra time taken by the second order methods does not adversely affect its performance. Thus, in the subsequent sections we compare the train loss and test accuracy versus time to evaluate the performance of the proposed method.

%oNAQ and oBFGS are slow and require longer time compared to the first order methods. This is  due to the Hessian computation and twice gradient calculation. The limited memory algorithms however require almost the same time as the first order methods.      % Further we analyse the amount of data accessed to attain a training error ${<10^{-3}}$. %which is illustrated in Fig.\ref{fig:eps} From Fig.\ref{fig:eps} we can see that the proposed algorithm can converge to a training error ${<10^{-3}}$ within fewer epochs compared to oBFGS, Adam and SGD. %Thus we can conclude that the proposed algorithm is faster.
%\vspace{-5mm}

\begin{figure}[h!]
  \centering
    \hspace{-3mm}
  \begin{subfigure}[b]{0.5\linewidth}
    \includegraphics[width=\linewidth]{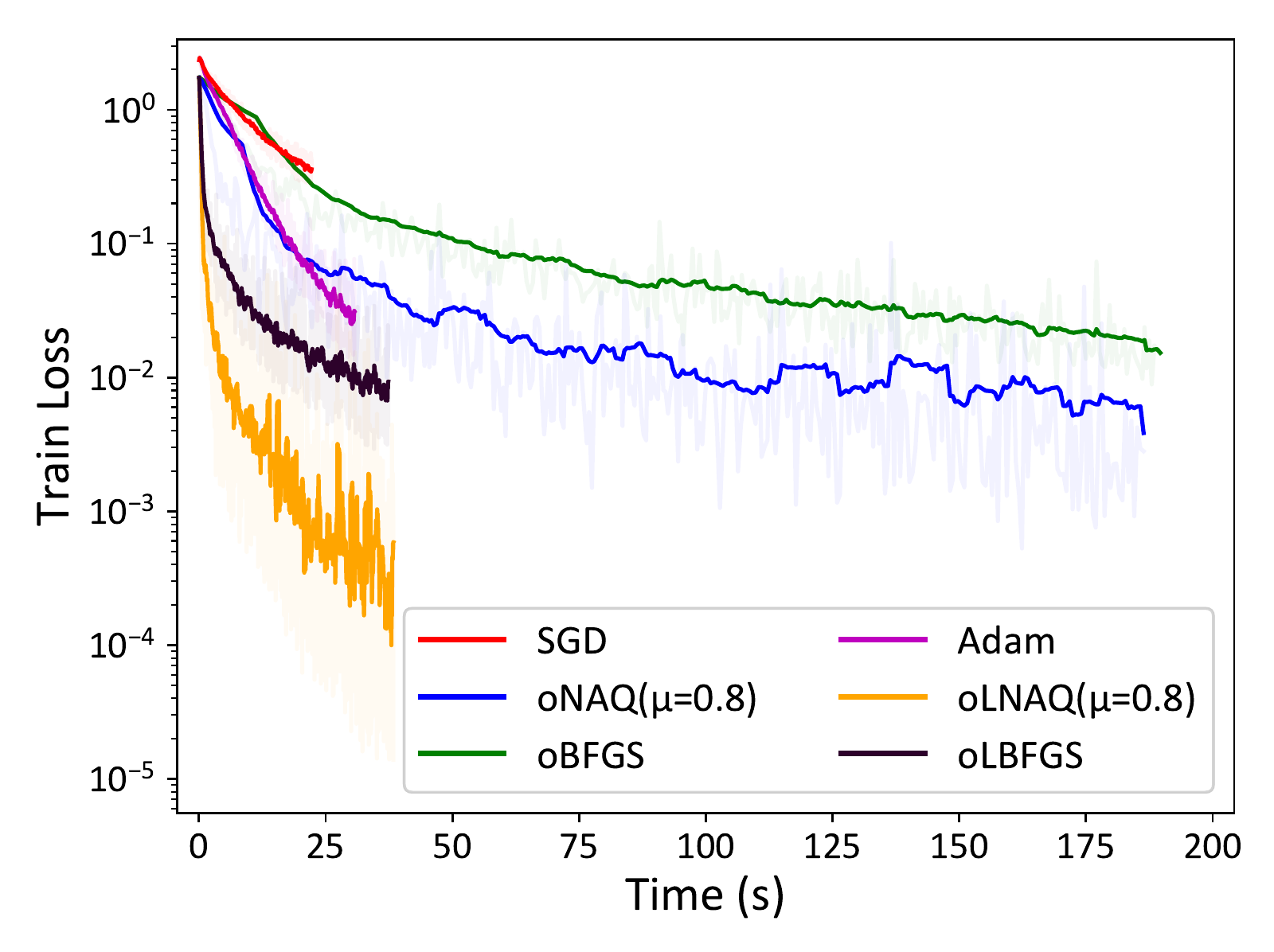}
    %\caption{Training Loss}
  \end{subfigure}
  \begin{subfigure}[b]{0.5\linewidth}
    \includegraphics[width=\linewidth]{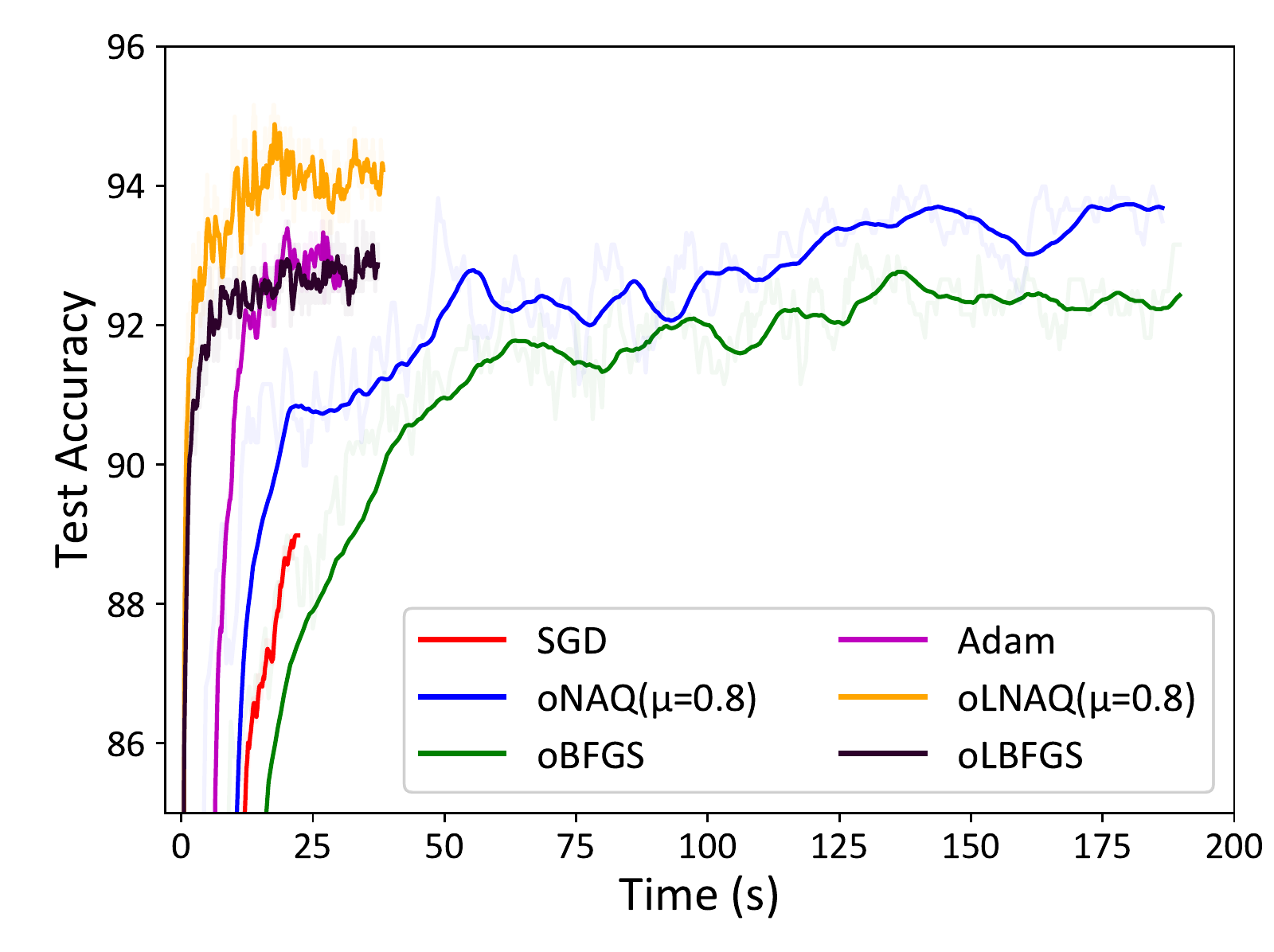}
    %\caption{Training Accuracy}
  \end{subfigure}
  %\caption{Time comparison of oLNAQ with ${b = 64}$ ${m = 4}$ and ${\mu = 0.8 }$ on 8x8 MNIST data }
  \caption{Comparison of train loss and test accuracy over time on 8x8 MNIST (80 epochs).}
  \label{fig:oNAQ}
\end{figure}

\begin{figure}[h!]
  \centering
  \hspace{-3mm}
  \begin{subfigure}[b]{0.5\linewidth}
    \includegraphics[width=\linewidth]{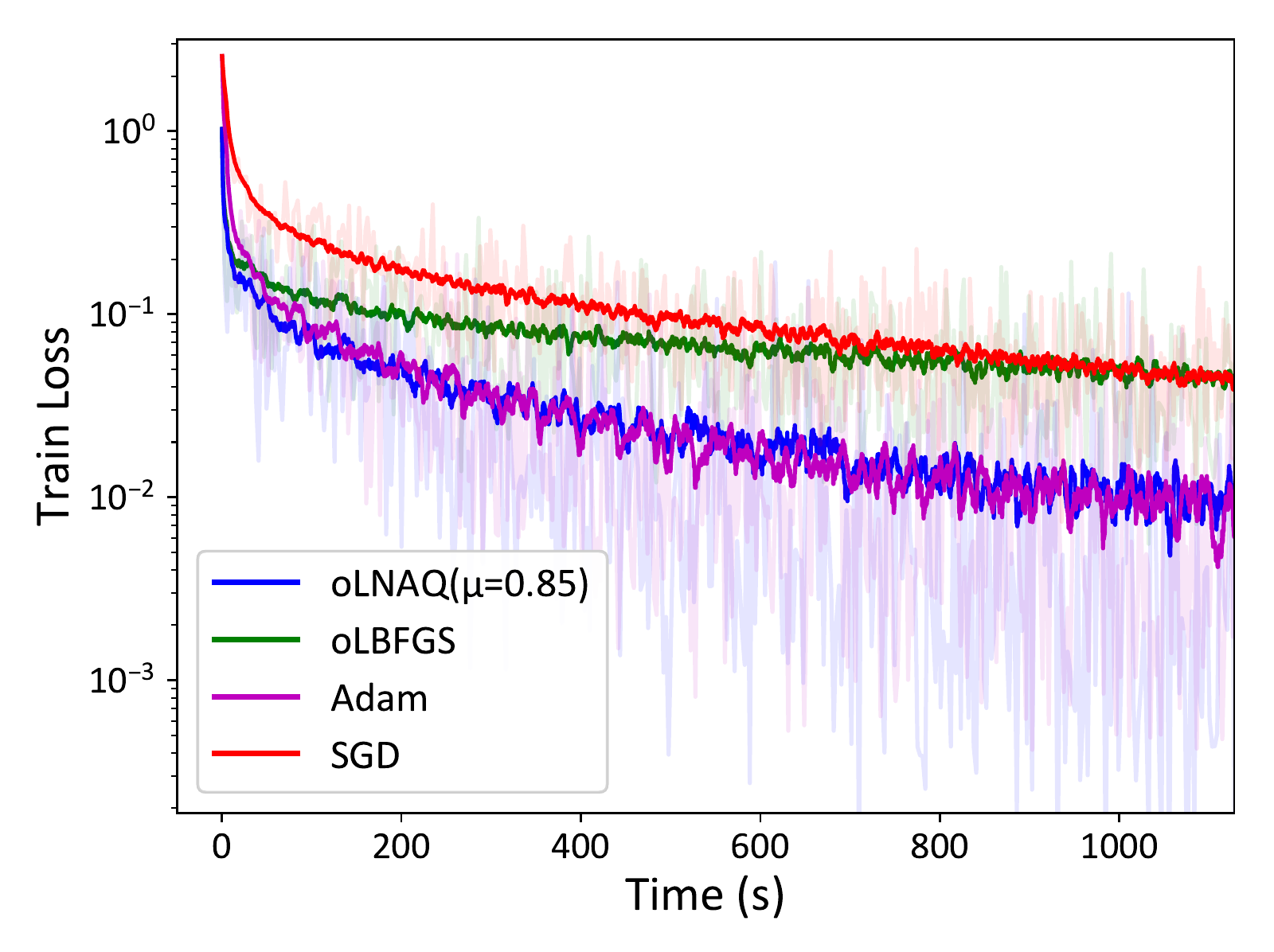}
    %\caption{Test Loss}
  \end{subfigure}
  %\hspace{-3mm}
  \begin{subfigure}[b]{0.5\linewidth}
    \includegraphics[width=\linewidth]{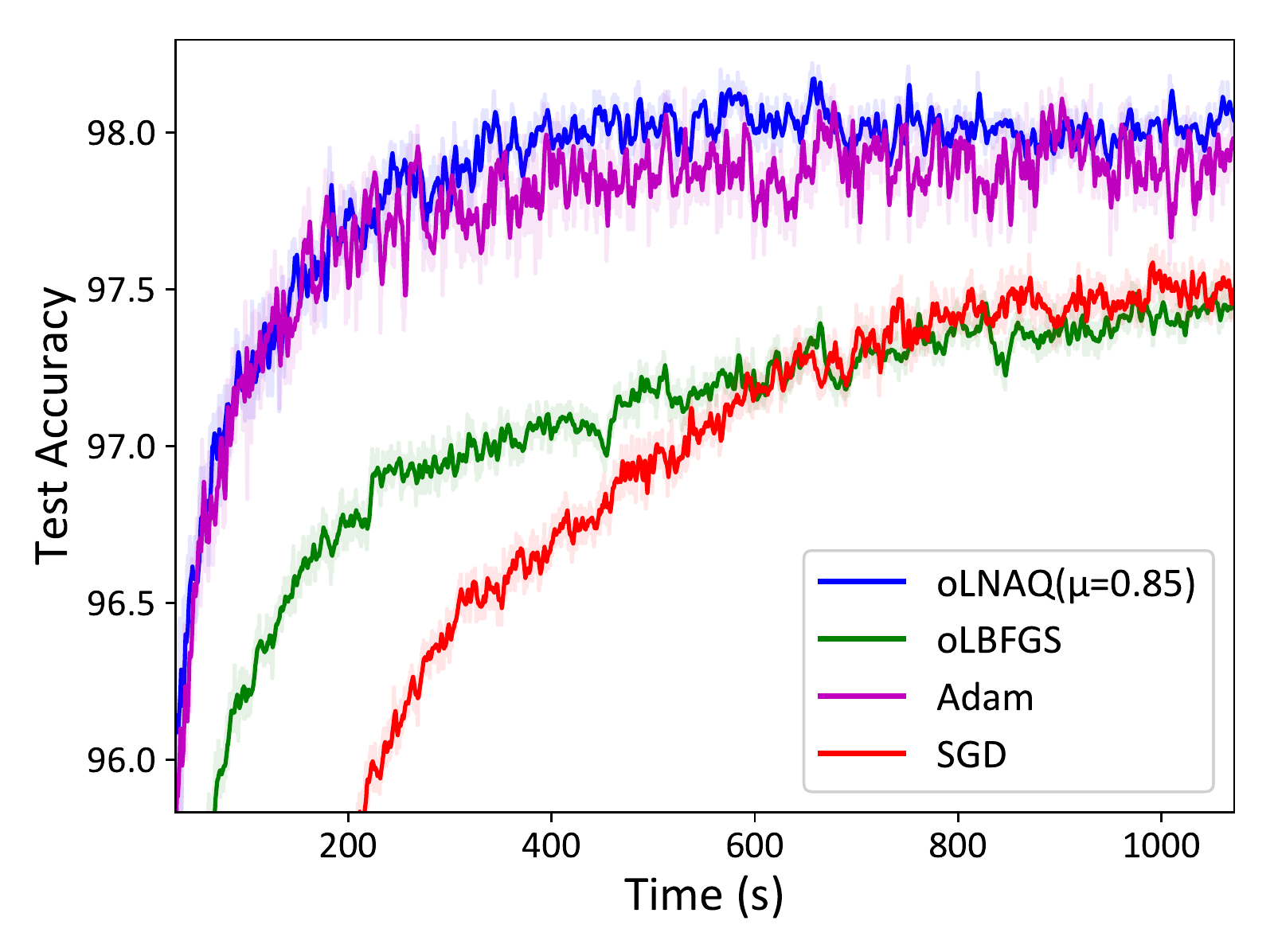}
    %\caption{Test Accuracy}
  \end{subfigure}
  
     \hspace{-3mm}
  \begin{subfigure}[b]{0.5\linewidth}
    \includegraphics[width=\linewidth]{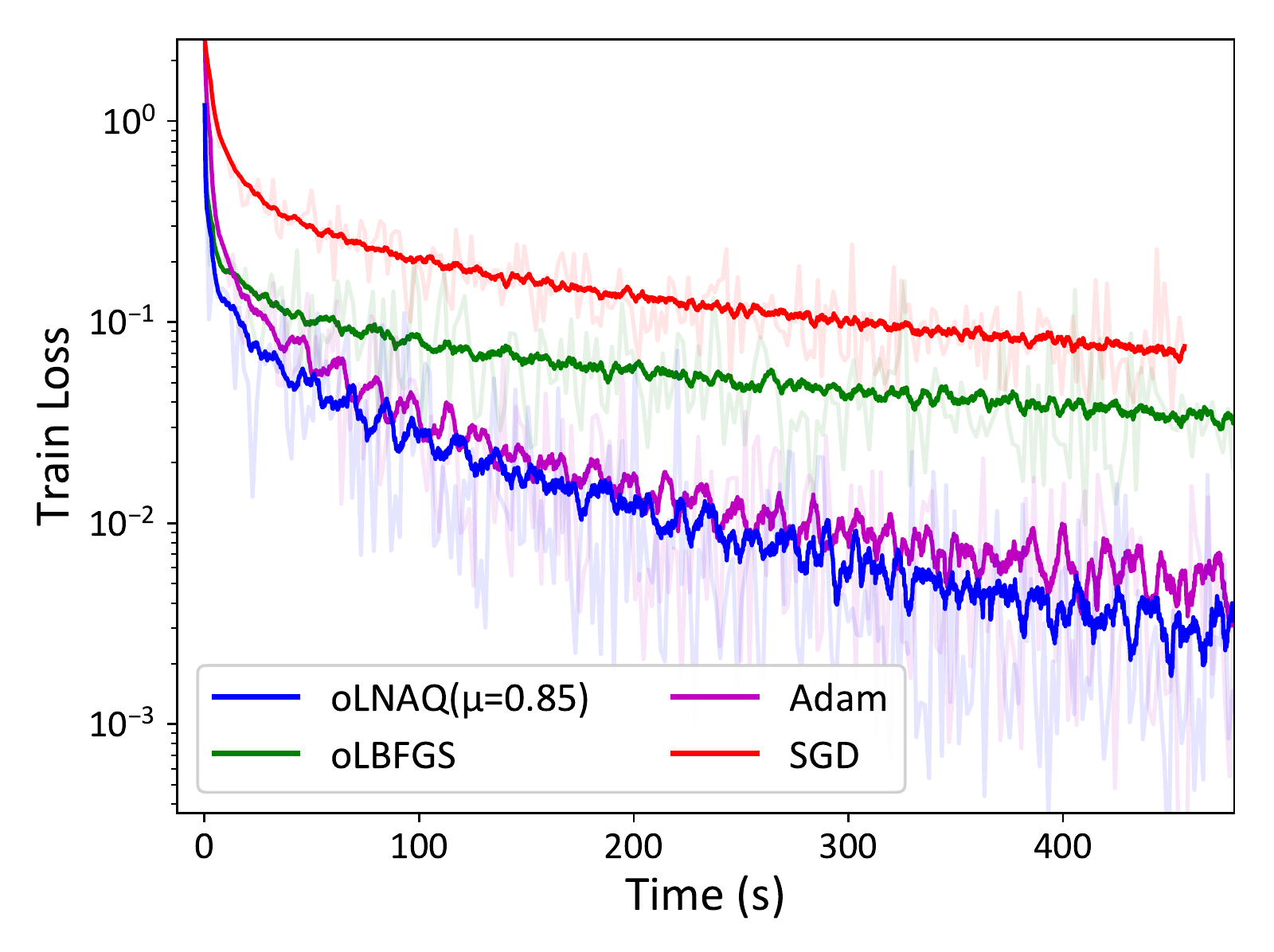}
    %\caption{}
  \end{subfigure}
  \begin{subfigure}[b]{0.5\linewidth}
    \includegraphics[width=\linewidth]{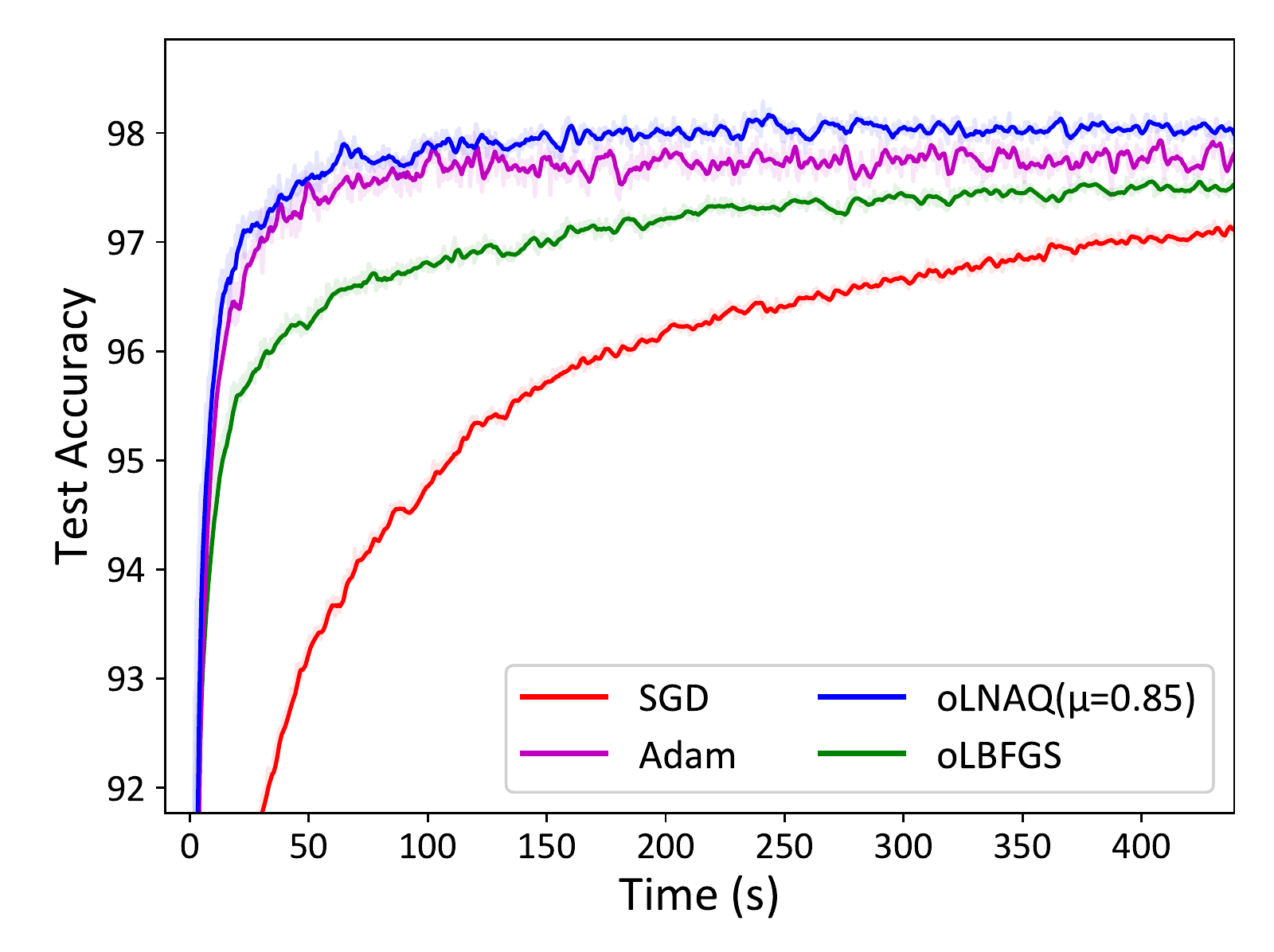}
    %\caption{}
  \end{subfigure}

  \caption{Results on 28x28 MNIST for ${b=64}$ (top) and ${b=128}$ (bottom).}
  \label{fig:oLNAQ}
\end{figure}

%\vspace{-8mm}
\subsubsection{Results on 28x28 MNIST Dataset}
Next, we evaluate the performance of the proposed algorithm on the standard 28x28 pixel MNIST dataset \cite{mnist2010data}. Due to system constraints and large number of parameters, we illustrate the performace of only the limited memory methods. %Fig.\ref{fig:batch} shows the performance comparison of oLNAQ and oLBFGS for different batch sizes ${b = 16,32,64,128}$. Fig.\ref{fig:memory} shows the performance for different memory sizes ${m = 4,8,10,16}$ with batch size 64 and ${\mu=0.8}$. We can see that oLNAQ performs well, giving a test accuracy close to 96.5\% even with small memory and batch sizes.   
%We compare the time taken by each algorithm to converge to a training error of ${10^{-2}}$. oNAQ and oLNAQ compute the gradient twice per iteration which increases the time per iteration. However this is compensated for since the algorithm can achieve lower error rates much faster compared to oLBFGS, Adam and SGD as seen in Fig.\ref{fig:time}. 
Fig.\ref{fig:oLNAQ} shows the results of oLNAQ %with a memory size of ${m=4}$ 
on the 28x28 MNIST dataset for batch size ${b=64}$ and ${b=128}$. %The momentum term $\mu = 0.85 $ is chosen. 
The results indicate that oLNAQ clearly outperforms oLBFGS and SGD for even small batch sizes. On comparing with Adam, oLNAQ is in close competition with Adam for small batch sizes such as ${b=64}$ and performs better for larger batch sizes such as ${b=128}$. %Increasing the memory from ${m=4}$ to ${m=8}$ or ${16}$ for a batch size ${b=64}$ improves the performance of oLNAQ when compared to Adam. %However considering computational efficiency, memory size is usually maintained smaller than the batch size. Since the computation cost is ${2bd+6md}$, if ${b\approx m}$ the computation cost would increase to ${8bd}$. Hence a smaller memory is desired.

%D:\Spring19\ECML\Figures\fig6_28x28\res28b64m4_784_100_50_10

\subsection{Convolution Neural Network - Classification Task }
We study the performance of the proposed algorithm on a simple convolution neural network (CNN) with two convolution layers followed by a fully connected layer.
%The structure of the CNN used is ${input-C_{5,3}-C_{5,5}-fc10-output}$ where ${C_{k,f}}$ denotes a convolution layer with ${f}$ filters of ${k}$x${k}$ kernel size.  %We evaluate the peformance of oNAQ on 8x8 MNIST dataset and oLNAQ on 28x28 MNIST dataset.The number of parameters ${d}$ is 778 and 3028 respectively. 
We use sigmoid activation functions and softmax cross-entropy error function. 
We evaluate the performance of oNAQ using the 8x8 MNIST dataset with a batch size of 64 and ${\mu = 0.8}$ and number of parameters ${d=778}$. The CNN architecture comprises of two convolution layers of 3 and 5 5x5 filters respectively, each followed by 2x2 max pooling layer with stride 2. The convolution layers are followed by a fully connected layer with 10 hidden neurons. %Fig x. shows the results of oNAQ on the simple CNN.
Fig. \ref{fig:CNN8x8} shows the CNN results of 8x8 MNIST. Calculation of the gradient twice per iteration increases the time per iteration when compared to the first order methods. However this is compensated well since the overall performance of the algorithm is much better compared to Adam and SGD. Also the number of epochs required to converge to low error and high accuracies is much lesser than the other algorithms. In other words, the same accuracy or error can be achieved with lesser amount of training data. % This is illustrated in Fig. \ref{fig:CNNeps}. 
 Further, we evaluate the performance of oLNAQ using the 28x28 MNIST dataset with batch size ${b = 128 , m = 4}$ and ${d=260,068}$. The CNN architecture is similar to that as described above except that the fully connected layer has 100 hidden neurons. Fig.\ref{fig:CNN} shows the results of oLNAQ on the simple CNN. The CNN results show similar performance as that of the results on multi-layer neural network where oLNAQ outperforms SGD and oBFGS. Comparing with Adam, oLNAQ is much faster in the first few epochs and becomes closely competitive to Adam as the number of epochs increases.% Fig. \ref{fig:CNN} (bottom) shows the train loss and test accuracy of the first epoch.        
 
% Fig x. and Fig x. illustrate the time taken and number of epochs required to , We choose a batch size of ${b=128}$ and memory ${m=4}$. Fig.\ref{fig:CNN} shows the results on the proposed algorithm in comparison with Adam, SGD and oLBFGS on the simple CNN.
%\vspace{-2mm}

\subsection{Multi-layer Neural Network - Regression Problem }
We further extend to study the performance of the proposed stochastic methods on regression problems. For this task, we choose two benchmark datasets - prediction of white wine quality \cite{cortez2009modeling} and CASP \cite{rana2013physicochemical} dataset. We evaluate the performance of oNAQ and oLNAQ on multi-layer neural network as shown in Table \ref{tab2}. Sigmoid activation function and mean squared error (MSE) function is used. Each layer except the output layer is batch normalized. Both datasets were z-normalized to have zero mean and unit variance.

\subsubsection{Results on Wine Quality Dataset }

We evaluate the performance of oNAQ and oLNAQ on the Wine Quality\cite{cortez2009modeling} dataset to predict the quality of the white wine on a scale of 3 to 9 based on 11 physiochemical test values. We split the dataset in 80-20 \% for train and test set.  For the regression problems, oNAQ with smaller values of momemtum $\mu=0.8$ and $\mu=0.85$ show similar performance as that of oBFGS. Larger values of momentum resulted in better performance. Hence we choose a value of $\mu=0.95$ which shows faster convergence compared to the other methods. Further comparing the performance for different batch sizes, we observe that for smaller batch sizes such as $b=32$, oNAQ is close in performance with Adam and oLNAQ is initially fast and gradually becomes close to Adam. For bigger batch sizes such as $b=64$, oNAQ and oLNAQ are faster in convergence initially. Over time, oLNAQ continues to result in lower error while oNAQ gradually becomes close to Adam. Fig. \ref{fig:wine} shows the root mean squared error (RMSE) versus time for batch sizes ${b=32}$ and ${b=64}$.

 \begin{figure}[h!]
  \centering
  \hspace{-3mm}
  \begin{subfigure}[b]{0.5\linewidth}
    \includegraphics[width=\linewidth]{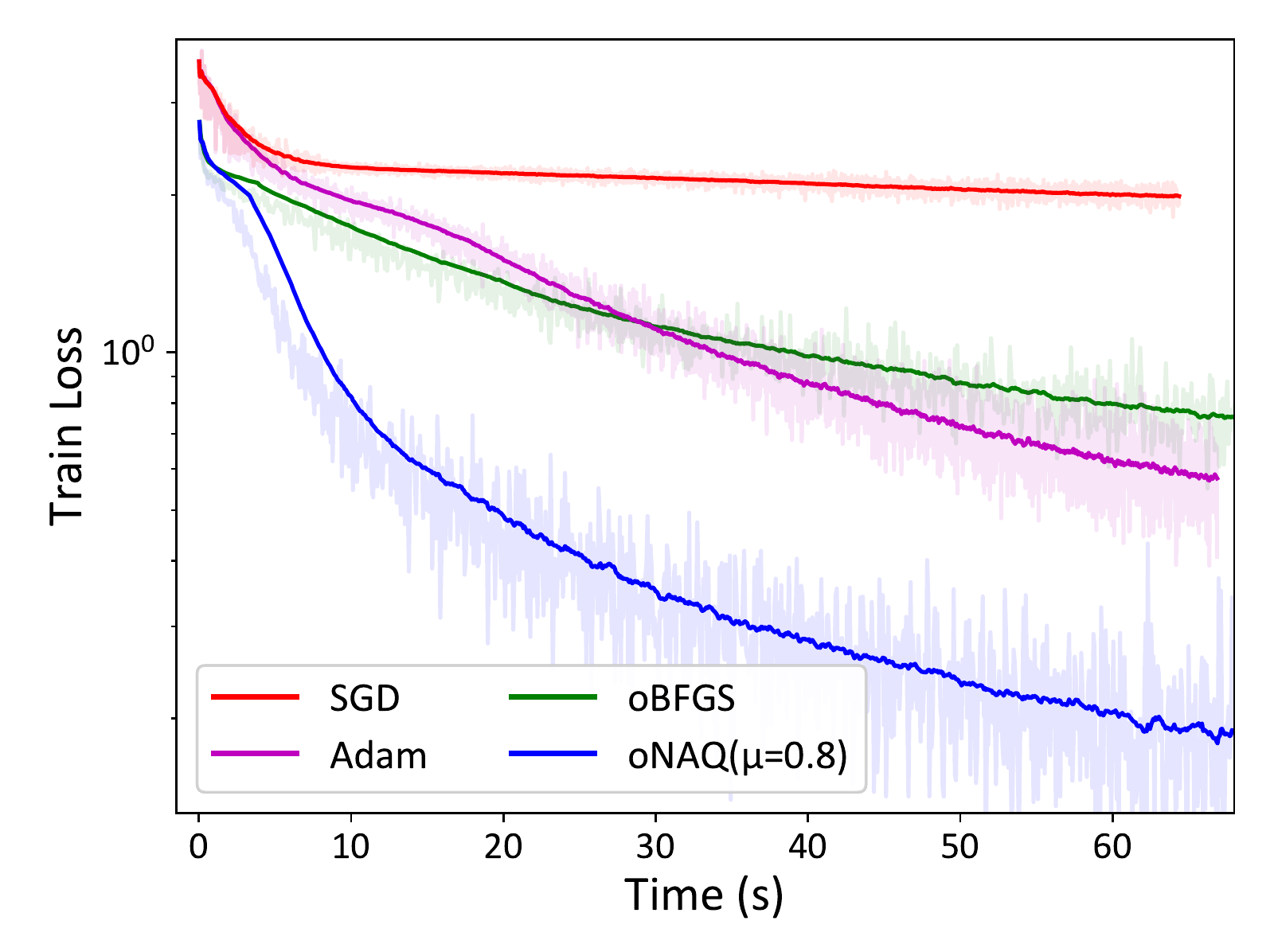}%8cnnTRL.png}
    %\caption{}
  \end{subfigure}
  \begin{subfigure}[b]{0.5\linewidth}
    \includegraphics[width=\linewidth]{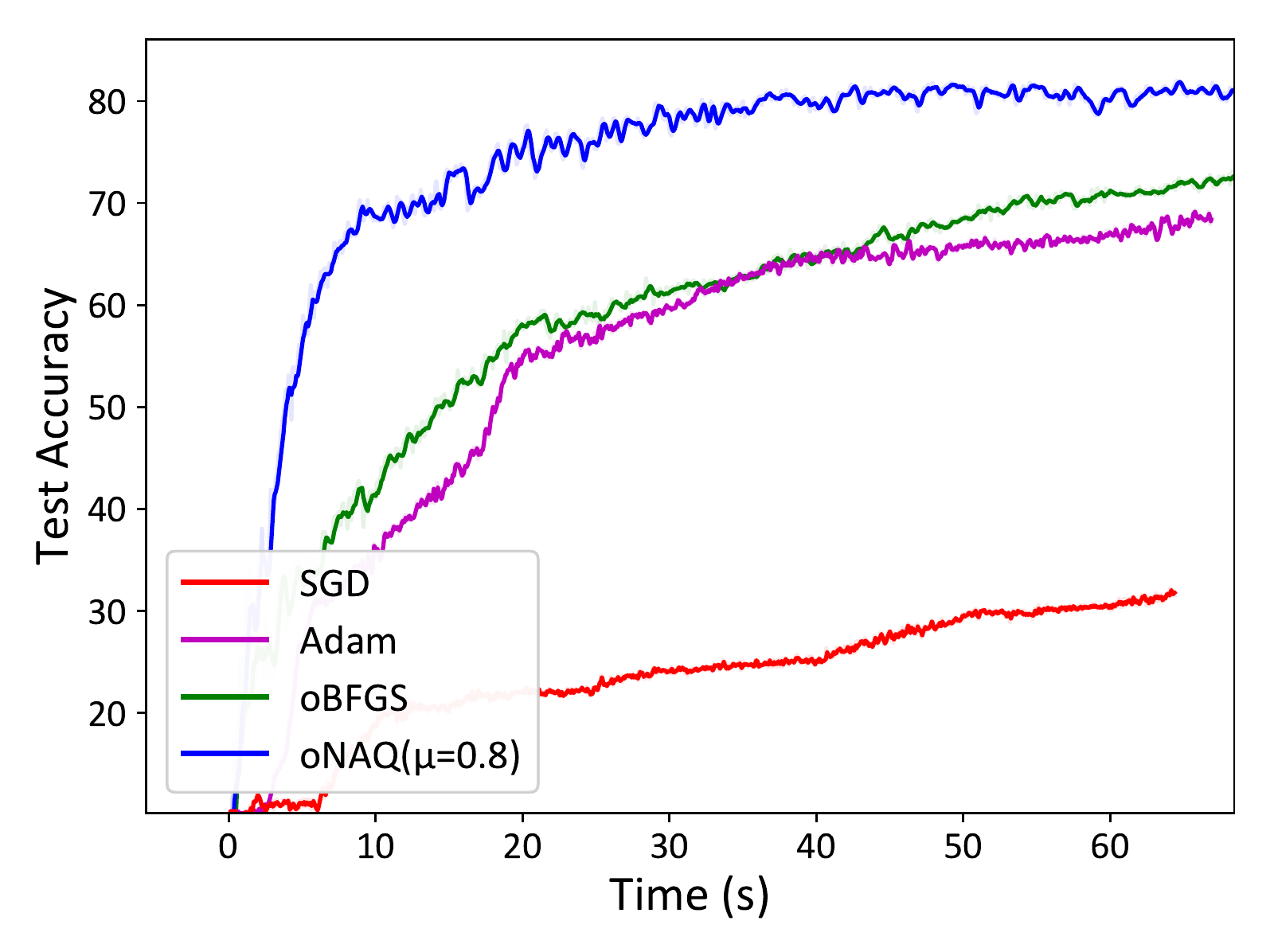}%8cnnTRA.png}
    %\caption{}
  \end{subfigure}
  \caption{ Convolution Neural Network results on 8x8 MNIST with ${b = 64}$.}%Error and Accuracy vs Data points accessed (epochs)  }
  \label{fig:CNN8x8}
  %\vspace{-5mm}
\end{figure}

\begin{figure}[h!]
  \centering
  \hspace{-3mm}
   \begin{subfigure}[b]{0.5\linewidth}
    \includegraphics[width=\linewidth]{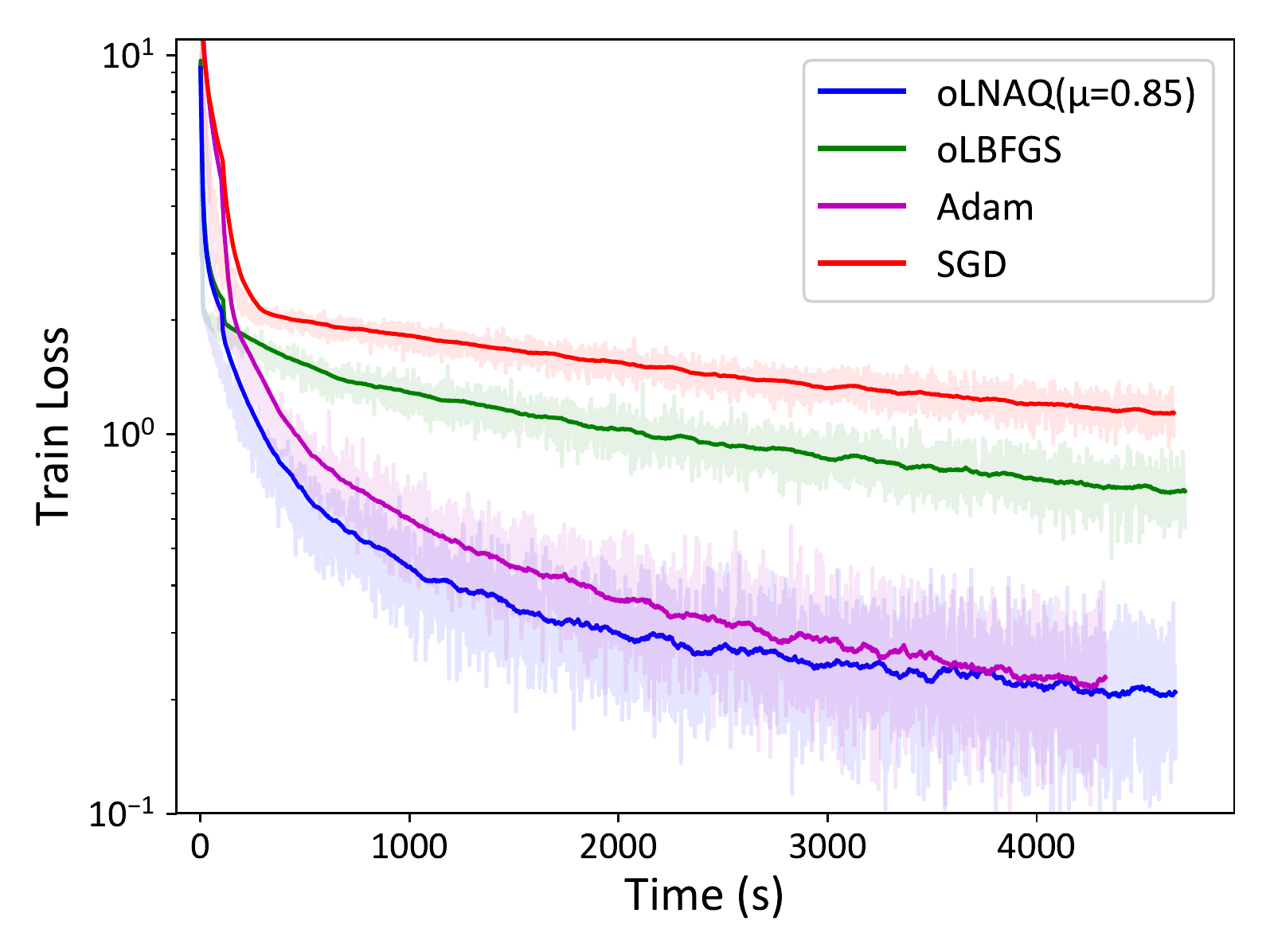}
    %\caption{Training Loss}
    \end{subfigure}
  \hspace{-3mm}
  \begin{subfigure}[b]{0.5\linewidth}
    \includegraphics[width=\linewidth]{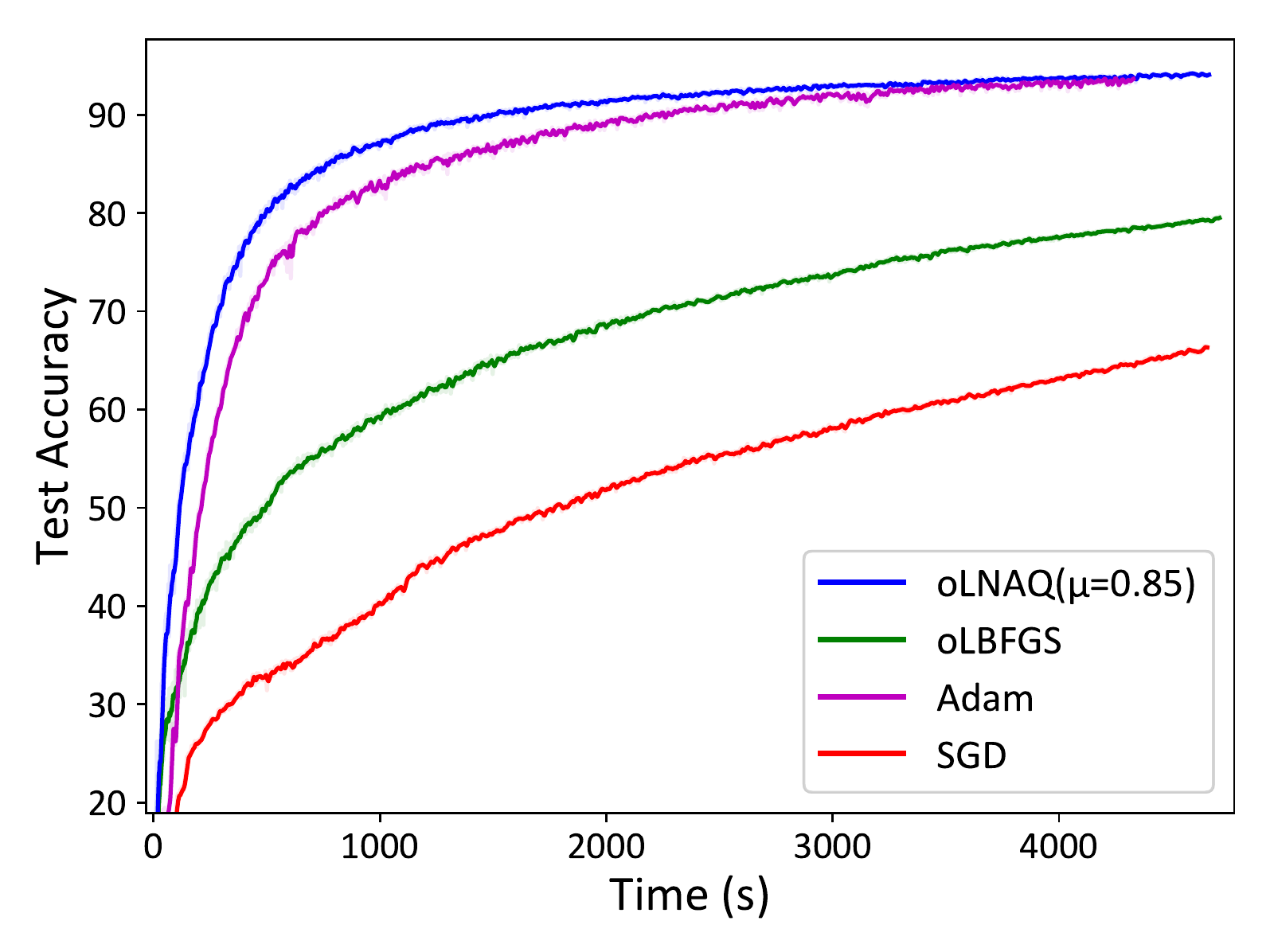}
    %\caption{Training Accuracy}
  \end{subfigure}
  
  \caption{CNN Results on 28x28 MNIST with ${b = 128}$. }
  \label{fig:CNN}
  %\vspace{-3mm}
\end{figure}

%\vspace{-2mm}
\begin{figure}[h!]
  \centering
  \hspace{-3mm}
  \begin{subfigure}[b]{0.5\linewidth}
    \includegraphics[width=\linewidth]{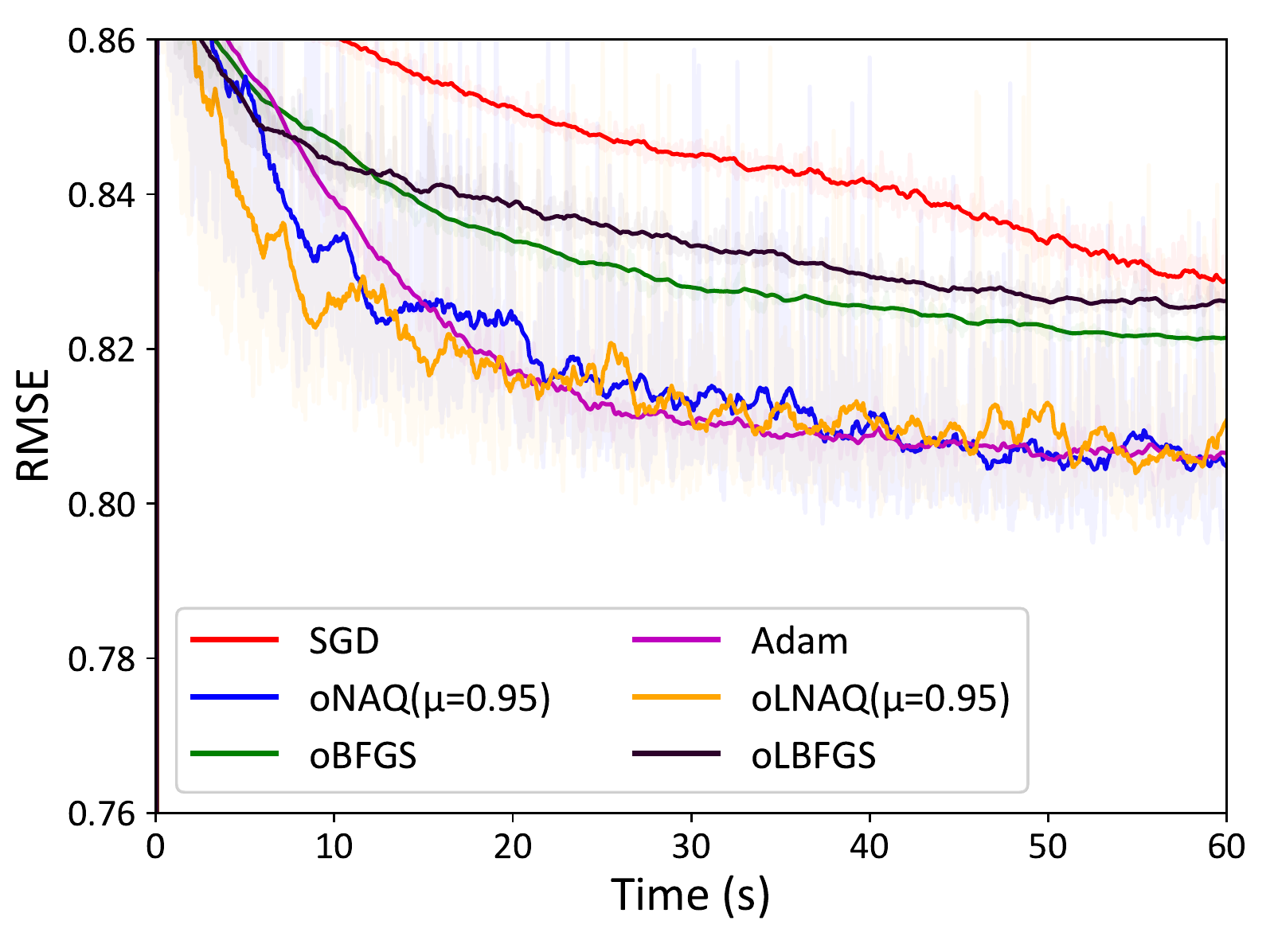}%{D:/Spring19/ECML/Figures/wine32/TRL_time.pdf}%fig10c.png
    %\caption{Test Loss}
  \end{subfigure}
  \hspace{-3mm}
  \begin{subfigure}[b]{0.5\linewidth}
    \includegraphics[width=\linewidth]{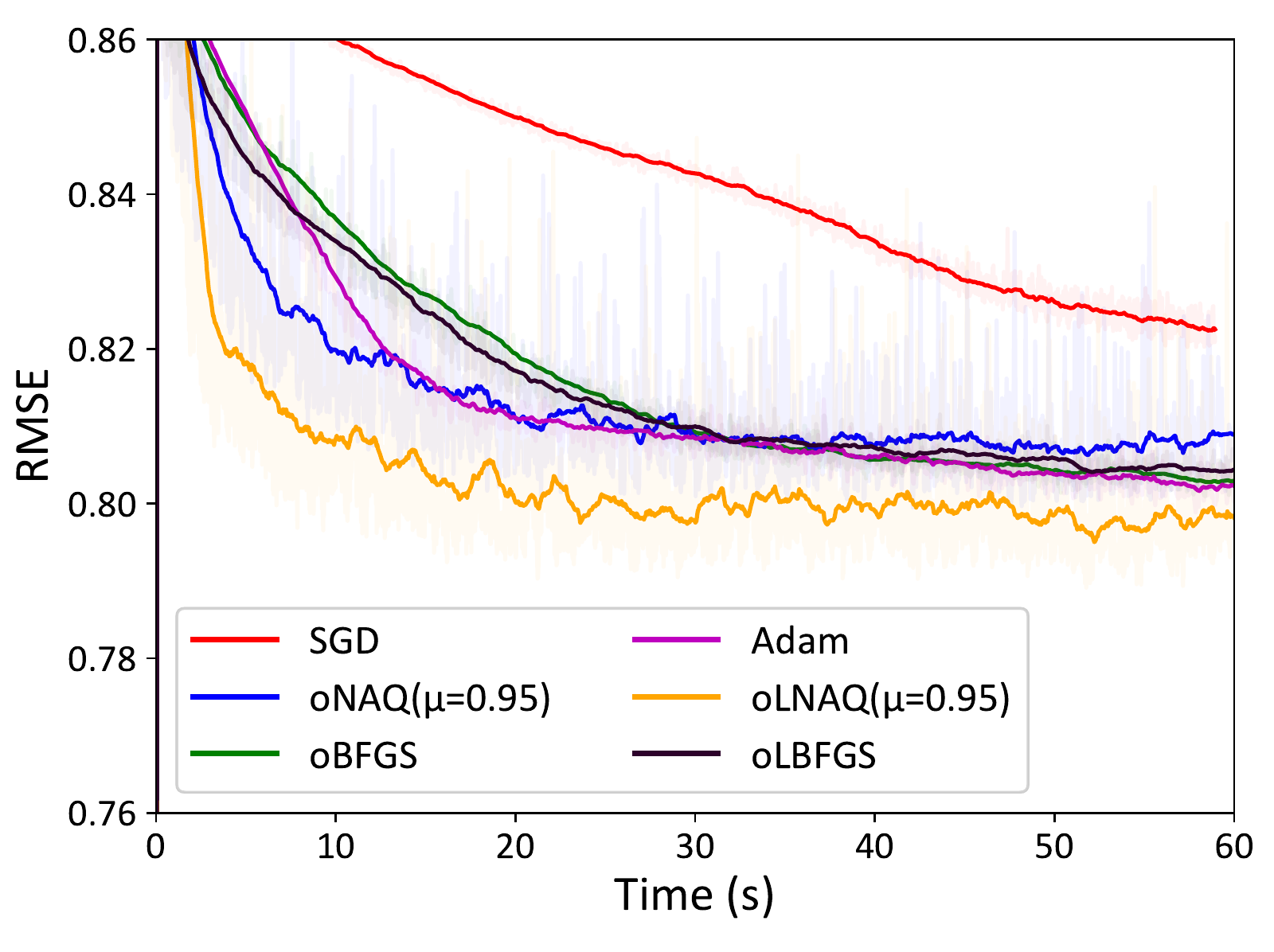}
    %\caption{Test Accuracy}
  \end{subfigure}
  \caption{Results of Wine Quality Dataset for ${b=32}$ (left) and ${b=64}$ (right). }
  \label{fig:wine}
\end{figure}

%\vspace{-15mm}
\begin{figure}[h!]
  \centering
  \hspace{-3mm}
  \begin{subfigure}[b]{0.5\linewidth}
    \includegraphics[width=\linewidth]{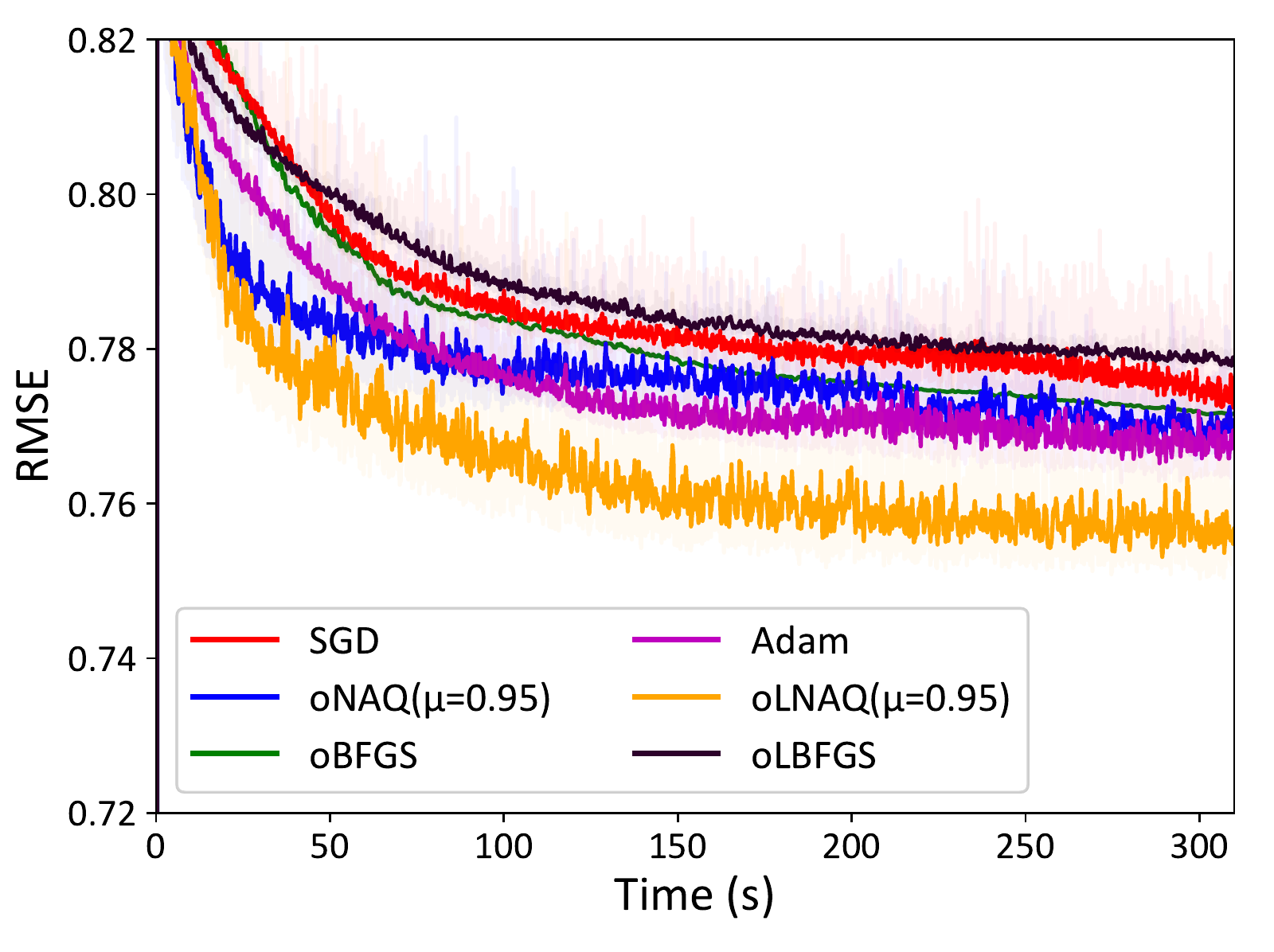}
    %\caption{Test Loss}
  \end{subfigure}
  \hspace{-3mm}
  \begin{subfigure}[b]{0.5\linewidth}
    \includegraphics[width=\linewidth]{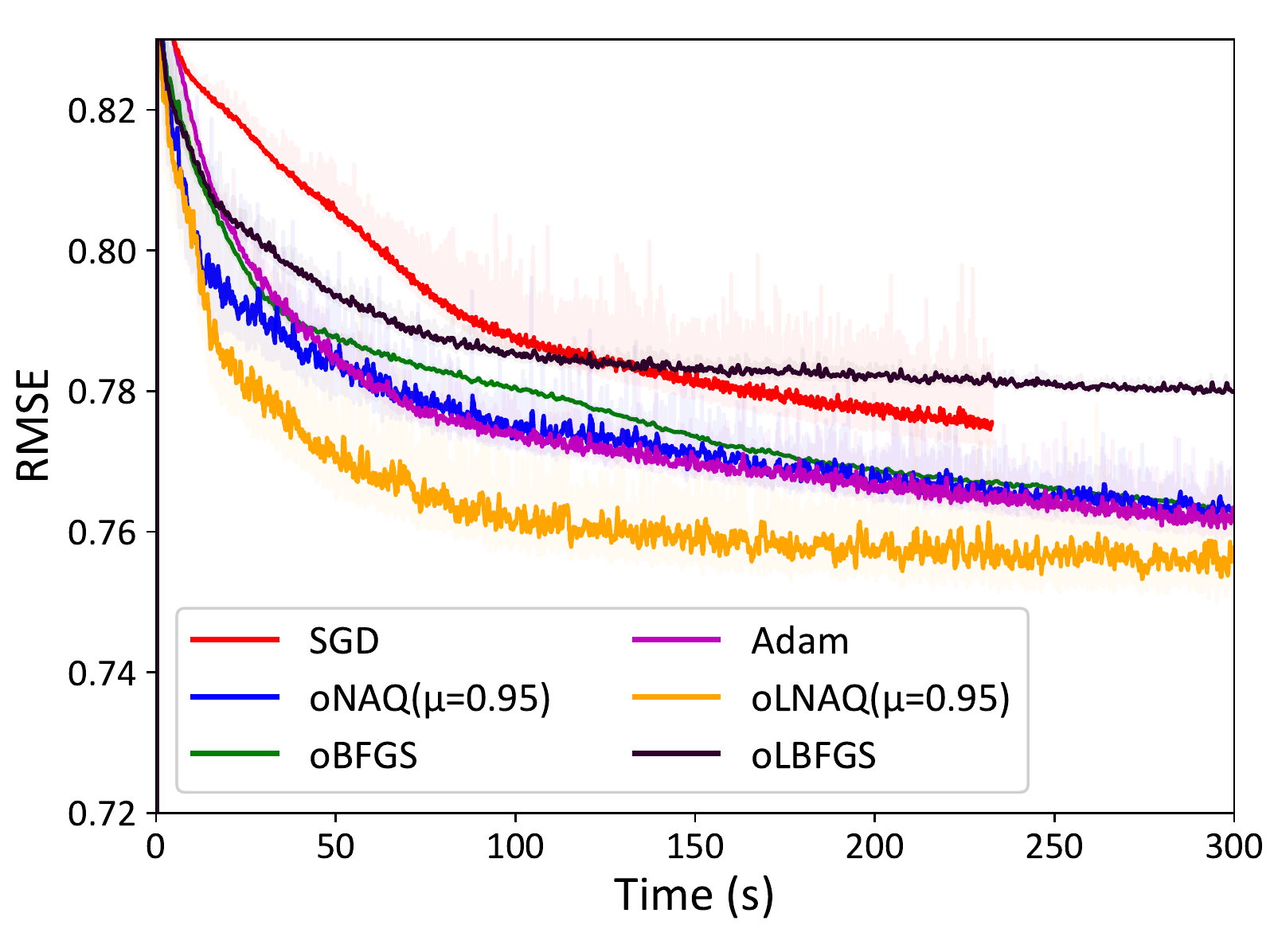}
    %\caption{Test Accuracy}
  \end{subfigure}
  \caption{Results of CASP Dataset for batch size ${b=64}$ (left) and ${b=128}$ (right).}
  \label{fig:casp}
  %\vspace{-10mm}
\end{figure}

\begin{figure}
\centering
\begin{minipage}{.5\textwidth}
  \centering
  \hspace{-3mm}
  %\captionsetup{justification=centering}
  \captionsetup{width=0.95\linewidth}
  \includegraphics[width=1\linewidth]{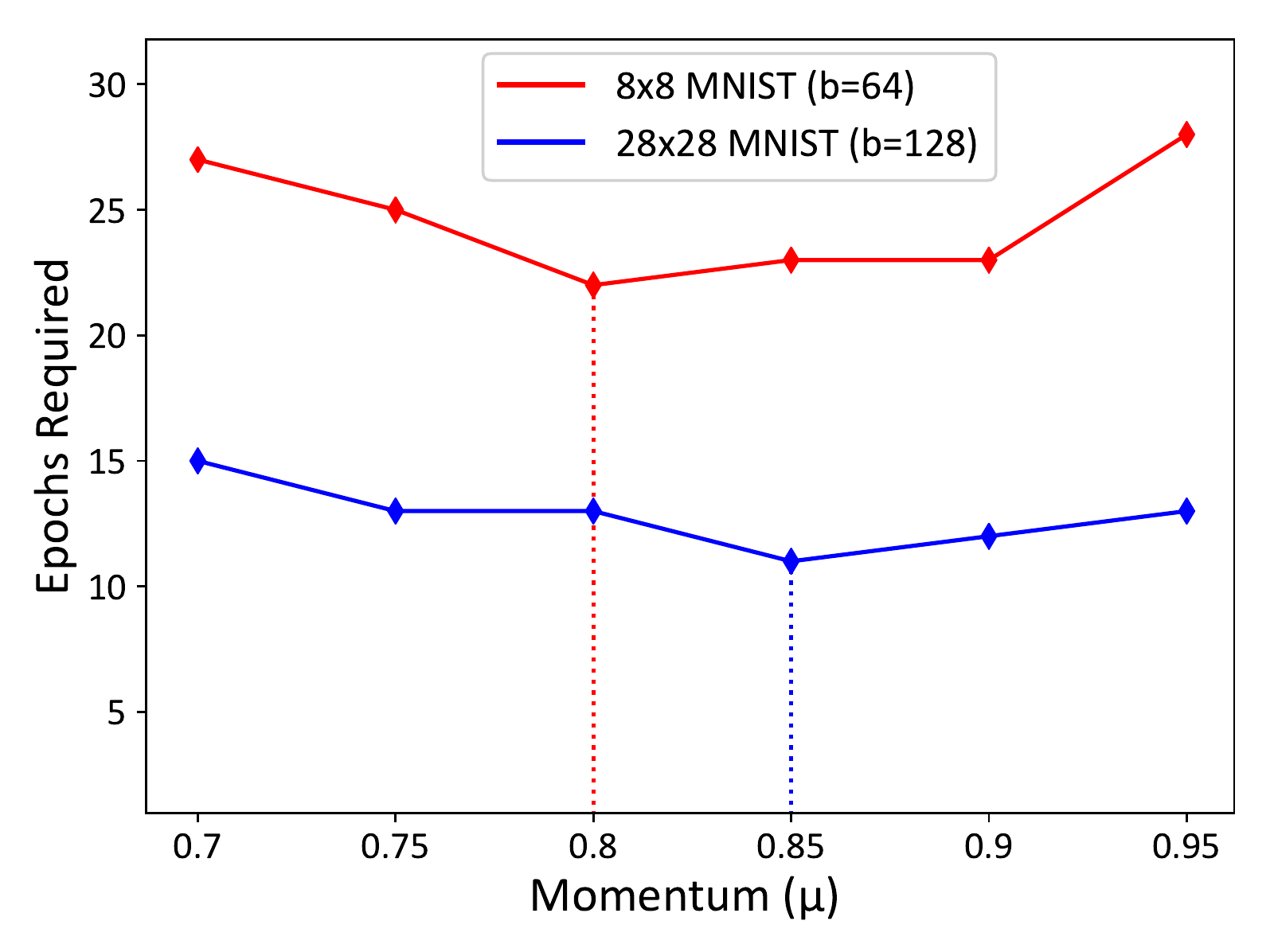}
  \captionof{figure}{No. of epochs required to converge for different values of ${\mu}$ with ${m=4}$ for oLNAQ classification problems.}
  %{Effect of momentum ${\mu}$ and batch size ${b}$ to converge to ${E({\bf w})<10^{-3}}$ for oNAQ algorithm with 8x8 MNIST}
  \label{fig:muoNAQ}
\end{minipage}%
\begin{minipage}{0.5\textwidth}
  \centering
  \hspace{-3mm}
  %\captionsetup{justification=centering}
   \captionsetup{width=0.9\linewidth}
  \includegraphics[width=1\linewidth]{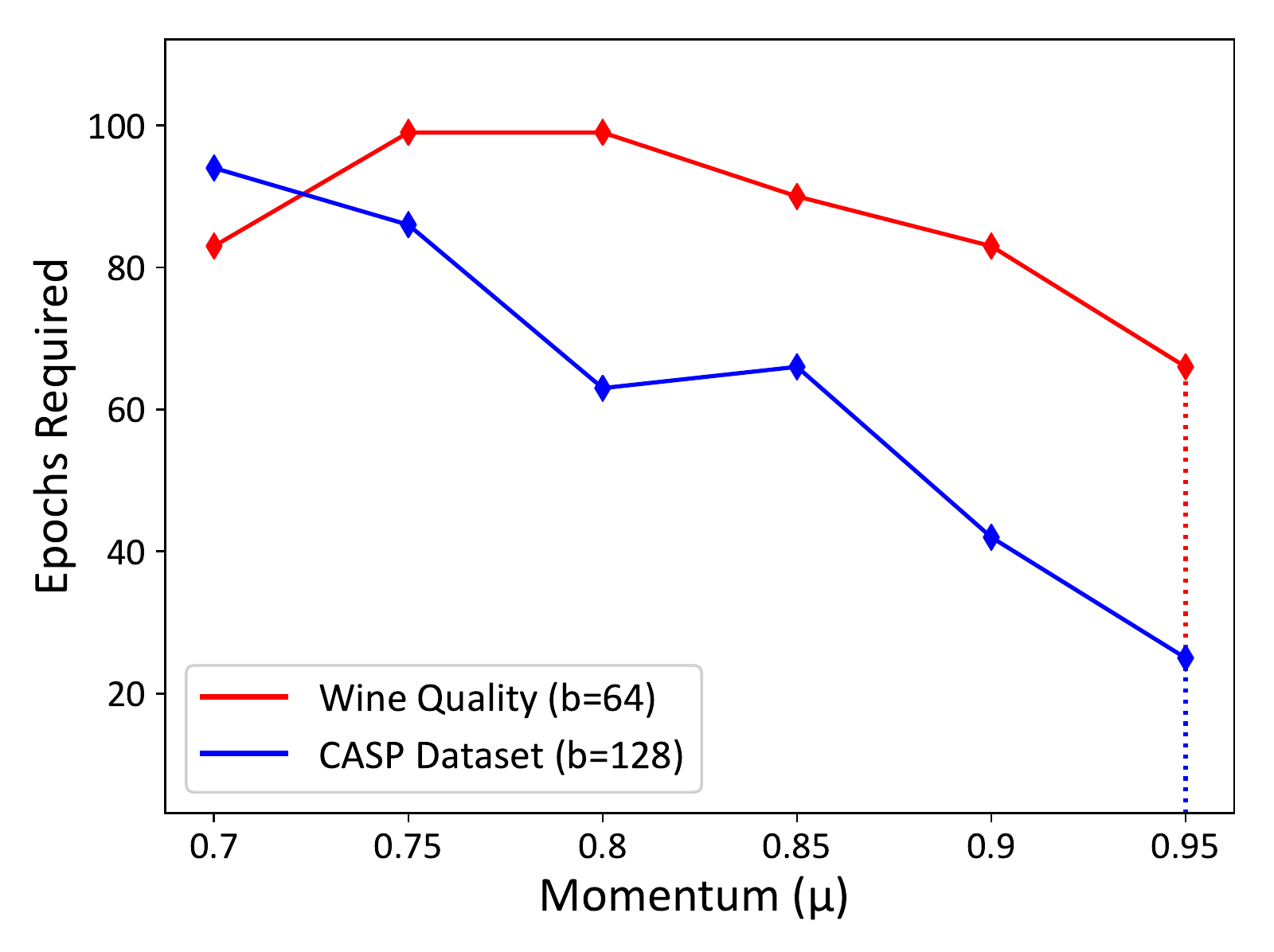}
  \captionof{figure}{No. of epochs required to converge for different values of ${\mu}$ with ${m=4}$ for oLNAQ regression problems.}
  \label{fig:muoLNAQ}
\end{minipage}
%\vspace{-2mm}
\end{figure}

%\vspace{-2mm}
\subsubsection{Results on CASP Dataset}
The next regression problem under consideration is the CASP (Critical Assessment of protein Structure Prediction) dataset from \cite{rana2013physicochemical}. It gives the  physicochemical properties of protein tertiary structure. We split the dataset in 80-20\% for train and test set. Similar to the wine quality problem, a momentum of ${\mu=0.95}$ was fixed. Fig. \ref{fig:casp} shows the root mean squared error (RMSE) versus time for batch sizes ${b=64}$ and ${b=128}$. %For both batch sizes oLNAQ shows consistent decrease in error much faster than the oBFGS, oLBFGS, SGD and Adam.
For both batch sizes, oNAQ in initially fast and becomes close to Adam and shows better performance compared to oBFGS and oLBFGS.  On the other hand, we observe that oLNAQ consistently shows decrease in error and outperforms the other algorithms for both batch sizes.

%The next regression problem under consideration is the CASP (Critical Assessment of protein Structure Prediction) dataset from \cite{nr}. It gives the  physicochemical properties of protein tertiary structure. We split the dataset in 80-20\% for train and test set. Similar to the wine quality problem, a momentum of ${\mu=0.95}$ was fixed. The simulation results for different batch sizes show that for smaller batch sizes, oLNAQ was faster initally and gradually remained constant while oNAQ continues to give lower error. For bigger batch sizes, oLNAQ consistently showed decrease in error much faster compared to oNAQ, oBFGS, oLBFGS, Adam and SGD.   Fig. \ref{fig:casp} shows the root mean squared error (RMSE) versus time for batch sizes ${b=64}$ and ${b=128}$. For both batch sizes, we observe that the oNAQ and oLNAQ outperforms the other algorithms.

%\vspace{-4mm}
\begin{table}
\begin{center}
\caption{Summary of Computational Cost and Storage. }\label{tab1}
\begin{tabular}{c c c c c l c c c}
\hline
\\[-3.2mm]
&&&&&{\;\;\;} & {\;\; Algorithm \;\;} &  {\;\;\;\;\; Computational Cost \;\;\;\;\;} & {\;\;\;\;\; Storage \;\;\;\;\;}\\
\hline
\\[-2mm]
&&&&\multirow{4}{*}{\rotatebox{90}{full batch}} & 
&{BFGS} &  ${nd + d^2 + \zeta nd}$ & ${d^2}$\\
&&&&&&{NAQ} &  ${2nd + d^2 + \zeta nd}$ & ${d^2}$\\
\\[-3mm]
&&&&&&{ LBFGS} &  ${nd + 4md + 2d + \zeta nd}$ & ${2md}$\\
&&&&&&{LNAQ} &  ${2nd + 4md + 2d + \zeta nd}$ & ${2md}$\\

\\[-3mm]
\hline
\\[-2mm]
&&&&\multirow{5}{*}{\rotatebox{90}{online}} & 

{} &{oBFGS} &  ${2bd + d^2}$ & ${d^2}$\\
&&&&{} &&{oNAQ} &  ${2bd + d^2}$ & ${d^2}$\\
\\[-3mm]
&&&&{} &&{oLBFGS} &  ${2bd + 6md}$ & ${2md}$\\
&&&&{} && {oLNAQ} &  ${2bd + 6md}$ & ${2md}$\\
\\[-3mm]
%&&&&{} &&{Adam} &  ${2bd}$ & ${2d}$\\
%\\[-3mm]
\hline
\end{tabular}
\end{center}
%\vspace{-5mm}
\end{table}

\subsection{Discussions on choice of parameters}

The momentum term ${\mu}$ is a hyperparameter with a value in the range ${0 < \mu < 1}$ and is usually chosen closer to 1\cite{sutskever2013importance,ninomiya2017novel}. %We study the performance of different values of the momentum term $\mu$ for different batch sizes. Here we illustrate the effect of the momentum parameter ${\mu}$ for different batch sizes on 8x8 MNIST dataset. Fig. \ref{fig:muoNAQ} shows the number of epochs required to converge to a train loss ${<10^{-3}}$ vs batch size for oNAQ. A momentum of ${\mu=0.8}$ and ${\mu=0.85}$ can be regarded as an optimum choice as it requires fewer epochs to converge for most of the batch sizes. However, ${\mu=0.9}$ gives better results for a larger batch size of ${b=256}$. A batch size of ${b=64}$ is an optimum choice as the number of epochs required to converge is almost the same as ${b=128}$ for momentum values ${\mu=0.8}$ and ${\mu=0.85}$. %Lesser the number of epochs to converge implies lesser amount of training required for convergence. 
The performance for different values of the momentum term have been studied for all the four problem sets in this paper. Fig. \ref{fig:muoNAQ} and Fig. \ref{fig:muoLNAQ} show the number of epochs required for convergence for different values of  $\mu$ for the classification and regression datasets respectively. %Similarly we study the variation of momentum ${\mu}$ and memory ${m}$ in oLNAQ for different batch sizes for each problem set. Based on these results, the performance of the proposed algorithms were verfied on the four datasets. 
For the limited memory schemes, a memory size of ${m=4}$ showed optimum results for all the four problem datasets with different batch sizes. Larger memory sizes also show good performance. However considering computational efficiency, memory size is usually maintained smaller than the batch size. Since the computation cost is ${2bd+6md}$, if ${b\approx m}$ the computation cost would increase to ${8bd}$. Hence a smaller memory is desired. Memory sizes less than ${m=4}$ does not perform well for small batch sizes and hence ${m=4}$ was chosen. 
 % Further studies on the choice of batch size, memory size and momentum parameters have been included in the supplementary materials. 

%\vspace{-2mm}

\subsection{Computation and Storage Cost}
The summary of the computational cost and storage for full batch and stochastic (online) methods are illustrated in Table {\ref{tab1}}. 
The cost of function and gradient evaluations can be considered to be ${nd}$, where ${n}$ is the number of training samples involved and ${d}$ is the number of parameters. The Nesterov's Accelerated quasi-Newton (NAQ) method computes the gradient twice per iteration compared to the BFGS quasi-Newton method which computes the gradient only once per iteration. Thus NAQ has an additional ${nd}$ computation cost.  In both BFGS and NAQ algorithms, the step length is determined by line search methods which involves ${\zeta}$ function evaluations until the search condition is satisfied.  In the limited memory forms the Hessian update is approximated using the two-loop recursion scheme, which requires ${4md+2d}$ multiplications. In the stochastic setting, both oBFGS and oNAQ compute the gradient twice per iteration, making the compuational cost the same in both. Both methods do not use line search and due to smaller number of training samples (minibatch) in each iteration, the computational cost is smaller compared to full batch.  Further, in stochastic limited memory methods, an additional $2md$ evaluations are required to compute the search direction as given (\ref{eq:oLBFGS}). In stochastic methods the computational complexity is reduced significantly due to smaller batch sizes (${b < n}$).

\iffalse

\begin{table}
\begin{center}
\caption{Summary of Results}\label{tab2}
\begin{tabular}{c c c c c c}
\hline
\\[-3mm]
{\;\; Algorithm \;\;} & {\;\; Time \;\;}& {\;\;\; 8x8 MNIST \;\;\;}& {\;\;\; 28x28 MNIST \;\;\;}& {\;\;\; Wine Quality \;\;\;}& {\;\;\; CASP \;\;\;}\\
\hline
\\[-3mm]
{SGD} &  \\
{Adam} &  \\
{oBFGS} &  \\
{oNAQ} & {0.00349} &  \\
{oLBFGS} & {0.00504}  \\
{oLNAQ} &  \\
\hline
\end{tabular}
\end{center}
\end{table}
\fi

%\vspace{-5mm}
\section{Conclusion}

In this paper we have introduced a stochastic quasi-Newton method with Nesterov's accelerated gradient. The proposed algorithm is shown to be efficient compared to the state of the art algorithms such Adam and classical quasi-Newton methods. 
From the results presented above, we can conclude that the proposed o(L)NAQ methods clearly outperforms the conventional o(L)BFGS methods with both having the same computation and storage costs. However the computation time taken by oBFGS and oNAQ are much higher compared to the first order methods due to Hessian computation. On the other hand, we observe that the per iteration computation of Adam, oLBFGS and oLNAQ  are comparable. By tuning the momentum parameter ${\mu}$, oLNAQ is seen to perform better and faster compared to Adam.  Hence we can conclude that with an appropriate value of ${\mu}$, oLNAQ can achieve better results. Further,
% The proposed algorithm shows good performance even in case of small batches and hence can be suitable for real-time continuous stream applications. 
the limited memory form of the proposed algorithm can efficiently reduce the memory requirements and computational cost  while incorporating second order curvature information. %The effect of the parameters such as momemtum ${\mu}$, memory ${m}$ and batch size ${b}$ have been studied.oLNAQ not only performs better than its second order counterpart oLBFGS but is also quite close to the first order Adam. 
Another observation is that the proposed oNAQ and oLNAQ methods significantly accelerates the training especially in the first few epochs when compared to both, first order Adam and second order o(L)BFGS method. Several studies propose pretrained models. oNAQ and oLNAQ can possibly be suitable for pretraining. Also, the computational speeds of oNAQ could be improved further by approximations which we leave for future work. %Further, in this paper the performance of the proposed algorithm has been evaluated on small neural networks i.e. with fewer number of hidden neurons and hidden layers. %Although there is no specific reason for choosing the number of hidden neurons in each hidden layers, it was observed that the proposed algorithm gave good results even for a small network such as 784-100-50-10. 
Further studying the performance of the proposed algorithm on bigger problem sets, including that of convex problems and on popular NN architectures such as AlexNet, LeNet and ResNet could test the limits of the algorithm. Furthermore, theoretical analysis of the convergence properties of the proposed algorithms will also be studied in future works.

%\bibliographystyle{splncs04}
%\bibliography{mybib}

%\bibliography{references}{}

\begin{thebibliography}{10}

\bibitem{haykin2009neural}
Haykin, S.:
\newblock Neural Networks and Learning Machines. 3rd edn.
\newblock Pearson Prentice Hall, (2009)

\bibitem{bottou2004large}
Bottou, L., Cun, Y.L.:
\newblock Large scale online learning.
\newblock In: Advances in neural information processing systems. (2004)
  217--224

\bibitem{bottou2010large}
Bottou, L.:
\newblock Large-scale machine learning with stochastic gradient descent.
\newblock In: Proceedings of COMPSTAT'2010.
\newblock Springer (2010)  177--186

\bibitem{robbins1951stochastic}
Robbins, H., Monro, S.:
\newblock A stochastic approximation method.
\newblock The annals of mathematical statistics (1951)  400--407

\bibitem{peng2019accelerating}
Peng, X., Li, L., Wang, F.Y.:
\newblock Accelerating minibatch stochastic gradient descent using typicality
  sampling.
\newblock arXiv preprint arXiv:1903.04192 (2019)

\bibitem{johnson2013accelerating}
Johnson, R., Zhang, T.:
\newblock Accelerating stochastic gradient descent using predictive variance
  reduction.
\newblock In: Advances in neural information processing systems. (2013)
  315--323

\bibitem{duchi2011adaptive}
Duchi, J., Hazan, E., Singer, Y.:
\newblock Adaptive subgradient methods for online learning and stochastic
  optimization.
\newblock Journal of Machine Learning Research \textbf{12}(Jul) (2011)
  2121--2159

\bibitem{tieleman2012lecture}
Tieleman, T., Hinton, G.:
\newblock Lecture 6.5-rmsprop, coursera: Neural networks for machine learning.
\newblock University of Toronto, Technical Report (2012)

\bibitem{kingma2014adam}
Kingma, D.P., Ba, J.:
\newblock Adam : A method for stochastic optimization.
\newblock arXiv preprint arXiv:1412.6980 (2014)

\bibitem{martens2010deep}
Martens, J.:
\newblock Deep learning via hessian-free optimization.
\newblock In: ICML. Volume~27. (2010)  735--742

\bibitem{roosta2016sub}
Roosta-Khorasani, F., Mahoney, M.W.:
\newblock Sub-sampled newton methods i: globally convergent algorithms.
\newblock arXiv preprint arXiv:1601.04737 (2016)

\bibitem{dennis1977quasi}
Dennis, Jr, J.E., Mor{\'e}, J.J.:
\newblock Quasi-newton methods, motivation and theory.
\newblock SIAM review \textbf{19}(1) (1977)  46--89

\bibitem{schraudolph2007stochastic}
Schraudolph, N.N., Yu, J., G{\"u}nter, S.:
\newblock A stochastic quasi-newton method for online convex optimization.
\newblock In: Artificial Intelligence and Statistics. (2007)  436--443

\bibitem{mokhtari2014res}
Mokhtari, A., Ribeiro, A.:
\newblock Res: Regularized stochastic bfgs algorithm.
\newblock IEEE Transactions on Signal Processing \textbf{62}(23) (2014)
  6089--6104

\bibitem{mokhtari2015global}
Mokhtari, A., Ribeiro, A.:
\newblock Global convergence of online limited memory bfgs.
\newblock The Journal of Machine Learning Research \textbf{16}(1) (2015)
  3151--3181

\bibitem{byrd2016stochastic}
Byrd, R.H., Hansen, S.L., Nocedal, J., Singer, Y.:
\newblock A stochastic quasi-newton method for large-scale optimization.
\newblock SIAM Journal on Optimization \textbf{26}(2) (2016)  1008--1031

\bibitem{wang2017stochastic}
Wang, X., Ma, S., Goldfarb, D., Liu, W.:
\newblock Stochastic quasi-newton methods for nonconvex stochastic
  optimization.
\newblock SIAM Journal on Optimization \textbf{27}(2) (2017)  927--956

\bibitem{li2018implementation}
Li, Y., Liu, H.:
\newblock Implementation of stochastic quasi-newton's method in pytorch.
\newblock arXiv preprint arXiv:1805.02338 (2018)

\bibitem{lucchi2015variance}
Lucchi, A., McWilliams, B., Hofmann, T.:
\newblock A variance reduced stochastic newton method.
\newblock arXiv preprint arXiv:1503.08316 (2015)

\bibitem{moritz2016linearly}
Moritz, P., Nishihara, R., Jordan, M.:
\newblock A linearly-convergent stochastic l-bfgs algorithm.
\newblock In: Artificial Intelligence and Statistics. (2016)  249--258

\bibitem{bollapragada2018progressive}
Bollapragada, R., Mudigere, D., Nocedal, J., Shi, H.J.M., Tang, P.T.P.:
\newblock A progressive batching l-bfgs method for machine learning.
\newblock arXiv preprint arXiv:1802.05374 (2018)

\bibitem{byrd2011use}
Byrd, R.H., Chin, G.M., Neveitt, W., Nocedal, J.:
\newblock On the use of stochastic hessian information in optimization methods
  for machine learning.
\newblock SIAM Journal on Optimization \textbf{21}(3) (2011)  977--995

\bibitem{gower2016stochastic}
Gower, R., Goldfarb, D., Richt{\'a}rik, P.:
\newblock Stochastic block bfgs: Squeezing more curvature out of data.
\newblock In: International Conference on Machine Learning. (2016)  1869--1878

\bibitem{ninomiya2017novel}
Ninomiya, H.:
\newblock A novel quasi-newton-based optimization for neural network training
  incorporating nesterov's accelerated gradient.
\newblock Nonlinear Theory and Its Applications, IEICE \textbf{8}(4) (2017)
  289--301

\bibitem{LNAQ_shah}
Mahboubi, S., Ninomiya, H.:
\newblock A novel training algorithm based on limited-memory quasi-newton
  method with nesterov’s accelerated gradient in neural networks and its
  application to highly-nonlinear modeling of microwave circuit.
\newblock IARIA International Journal on Advances in Software \textbf{11}(3-4)
  (2018)  323--334

\bibitem{nocedal2006}
Nocedal, J., Wright, S.J.:
\newblock Numerical Optimization.
\newblock Springer Series in Operations Research. Springer, second edition
  (2006)

\bibitem{zhang2005globally}
Zhang, L.:
\newblock A globally convergent bfgs method for nonconvex minimization without
  line searches.
\newblock Optimization Methods and Software \textbf{20}(6) (2005)  737--747

\bibitem{dai2002convergence}
Dai, Y.H.:
\newblock Convergence properties of the bfgs algoritm.
\newblock SIAM Journal on Optimization \textbf{13}(3) (2002)  693--701

\bibitem{indrapriyadarsini2018implementation}
Indrapriyadarsini, S., Mahboubi, S., Ninomiya, H., Asai, H.:
\newblock Implementation of a modified nesterov's accelerated quasi-newton
  method on tensorflow.
\newblock In: 2018 17th IEEE International Conference on Machine Learning and
  Applications (ICMLA), IEEE (2018)  1147--1154

\bibitem{zinkevich2003online}
Zinkevich, M.:
\newblock Online convex programming and generalized infinitesimal gradient
  ascent.
\newblock In: Proceedings of the 20th International Conference on Machine
  Learning (ICML-03). (2003)  928--936

\bibitem{cortez2009modeling}
Cortez, P., Cerdeira, A., Almeida, F., Matos, T., Reis, J.:
\newblock Modeling wine preferences by data mining from physicochemical
  properties.
\newblock Decision Support Systems \textbf{47}(4) (2009)  547--553
  \url{https://archive.ics.uci.edu/ml/datasets/wine+quality}

\bibitem{rana2013physicochemical}
Rana, P.:
\newblock Physicochemical properties of protein tertiary structure data set.
\newblock UCI Machine Learning Repository (2013)
  \url{https://archive.ics.uci.edu/ml/datasets/Physicochemical+Properties+of+Protein+Tertiary+Structure}

\bibitem{alpaydin1998optical}
Alpaydin, E., Kaynak, C.:
\newblock Optical recognition of handwritten digits data set.
\newblock UCI Machine Learning Repository (1998)
  \url{https://archive.ics.uci.edu/ml/datasets/optical+recognition+of+handwritten+digits}

\bibitem{mnist2010data}
LeCun, Y., Cortes, C., Burges, C.:
\newblock Mnist handwritten digit database.
\newblock AT\&T Labs [Online] Available: \url{http://yann.lecun.com/exdb/mnist}
  (2010)

\bibitem{sutskever2013importance}
Sutskever, I., Martens, J., Dahl, G.E., Hinton, G.E.:
\newblock On the importance of initialization and momentum in deep learning.
\newblock ICML (3) \textbf{28}(1139-1147) (2013) ~5

\end{thebibliography}
%\bibliographystyle{splncs}
%\bibliographystyle{ieeetr}

\end{document}